%% file: emnlp2023.tex
\pdfoutput=1

\documentclass[11pt]{article}

\usepackage{EMNLP2023} 

\usepackage{times}
\usepackage{latexsym}
\usepackage{booktabs}
\usepackage{graphicx}
\usepackage[utf8]{inputenc}

\usepackage{xcolor}
\usepackage{listingsutf8}

\usepackage{xcolor}
\usepackage{appendix}

\graphicspath{ {./images/} }

\usepackage[T1]{fontenc}

\usepackage[utf8]{inputenc}

\usepackage{microtype}

\usepackage{inconsolata}

\usepackage{algorithm2e}
\usepackage{amsmath}
\usepackage{amsthm}
\theoremstyle{definition}

\usepackage{enumitem} 

\title{Language hooks: a modular framework for augmenting LLM reasoning that decouples tool usage from the model and its prompt}

\author{Damien de Mijolla, Wen Yang, Philippa Duckett, Christopher Frye, Mark Worrall  \\
Faculty Science Ltd \\
\texttt{damien.de-mijolla@faculty.ai}}
\begin{document}
\maketitle

\begin{abstract}
Prompting and fine-tuning have emerged as two competing paradigms for augmenting language models with new capabilities, such as the use of tools. Prompting approaches are quick to set up but rely on providing explicit demonstrations of each tool’s usage in the model’s prompt, thus coupling tool use to the task at hand and limiting generalisation. Fine-tuning removes the need for task-specific demonstrations of tool usage at runtime; however, this ties new capabilities to a single model, thus making already-heavier setup costs a recurring expense. In this paper, we introduce language hooks, a novel framework for augmenting language models with new capabilities that is decoupled both from the model’s task-specific prompt and from the model itself. The language hook algorithm interleaves text generation by the base model with the execution of modular programs that trigger conditionally based on the existing text and the available capabilities. Upon triggering, programs may call external tools, auxiliary language models (e.g.~using tool specific prompts), and modify the existing context. We benchmark our method against state-of-the-art baselines, find that it outperforms task-aware approaches, and demonstrate its ability to generalise to novel tasks.
\end{abstract}

\section{Introduction}

\input{./schematic}

A defining characteristic of human intelligence is the way we have externalised it into the world around us, from language and culture to the technology we have built and the knowledge we have accumulated. This externalisation of capabilities alleviates the need for humans to repeatedly solve problems of the past, allowing us to focus instead on novel aspects of the task at hand. 

In a similar vein, recent work has explored the augmentation of large language models (LLMs) with external tools to increase the range of their capabilities \cite{LMsFewShot, ScalingInstruction-FT-LMs, Chameleon}. One approach to this involves hard-coding the reasoning paths available to the model \cite{InterleavingRetrievalWithChainOfThought, DSP}, which is feasible when the logical flow of a task is fixed. Another requires demonstrations of tool usage in the model's task-specific prompt \cite{ReAct, PAL}. All such task-aware approaches broaden the model's capabilities but at the cost of narrowing its general applicability.

As an alternative, fine-tuning LLMs for tool use \cite{ToolkenGPT, Toolformer} embeds tool-awareness into the model itself, rather than filling its prompt or imposing upon its reasoning path. Such methods exhibit greater generalisability but come with their own drawbacks, namely in setup cost and in tightly coupling the new capabilities to a single specific model. 

Beyond tool usage, LLM capabilities can be augmented in many directions. For example, recent research has explored the possibility of detecting failure modes in language generation, e.g.~hallucination \cite{self-contradiction, DoLMsKnowWhenTheyHallucinate}. Whilst this research is encouraging, there are important use cases for which it is so-far insufficient, e.g.~for verifying age-restricted content in an LLM-powered app on a jurisdiction-specific basis.
Verification of the model's output in such cases should be independent of the base model, as well as transparent and customisable by the user.

In our view, LLMs should be augmented with new capabilities through an approach that is: 

\begin{enumerate}[noitemsep]
  \item \textit{Task-agnostic:} widely applicable to new problems and, in particular, independent from the model's task-specific prompt;
  \item \textit{Model-agnostic:} decoupled from any particular model to avoid recurrent fine-tuning costs and enable independent verification; 
  \item \textit{Modular:} built up from transparent components, each designed to enable a specific capability, with the option of combining any number of such components at runtime;
  \item \textit{Non-intrusive:} the model remains free to pursue any reasoning path and to use language naturally; e.g.~no specific control flow or text structure should be dictated by the method.
\end{enumerate}

In this paper, we introduce \emph{language hooks}\footnote{In computer programming, hooking refers to the interception and alteration of the behaviour of a software application.}, a novel paradigm for incorporating new capabilities into text generation with language models. Our procedure satisfies the key criteria laid out above. At its core, the language hook algorithm interleaves text generation by a base model with the execution of programs that may call external tools and modify the existing context. The event type (text generation or program execution) that occurs at each step in the algorithm is determined conditionally on the existing context, which consists exclusively of natural language. See Figure \ref{fig:schematic} for a schematic. This is a flexible framework for integrating systematic reasoning and tool use into text generation; it requires no model tuning and is compatible with any prompting strategy. 

Our contributions in this work are fourfold. First, we introduce a new framework for text generation with language hooks: small conditional programs that modify the existing context in between text generation events. Second, we provide concrete implementations of three specific hooks: for mathematical calculations, knowledge retrieval, and guardrail interception. Third, we benchmark language hooks against existing baselines, achieving competitive performance with approaches designed for narrower applications. Fourth, we demonstrate language hooks' applicability to more open-ended tasks with varied control flows and requiring the use of multiple tools. To our knowledge, this is the first work to propose an event-driven framework for multiple tool usage.

\section{Language hooks} \label{sec:language-hooks}

In this section, we define the language hook algorithm in detail and provide examples of specific hooks used in the experiments of Section \ref{experiments}.

We begin by defining notation. Let $C_0$ denote the prompt, e.g.~a question to be answered and any in-context examples. Let $g$ denote the base model performing text generation, which will proceed one sentence at a time. Let $s_t$ denote the $t$-th sentence of the response and $r_t = [s_1, \ldots, s_{t-1}, s_t]$ the reasoning stream. The full context after $t$ sentences is then $C_t = [C_0, r_{t-1}, s_t]$, and we have $C_{t+1} = g(C_t)$. We introduce $f: s_t \to \{0, 1\}$ to determine whether a stopping condition has been met, e.g.~whether $s_t$ contains ``Final answer:''.

\textbf{Definition.} We define a language hook (or ``hook'' for short) as a triplet $h=(p,\tau,e)$ consisting of a program $p$, a trigger $\tau$, and eligibility criteria $e$. These components will be defined in detail below, culminating in Algorithm \ref{alg:lh}, which is generally run with a set $H$ of several hooks.

\textbf{Program.} The program $p$ of a language hook is a function that takes the existing context $C_t$ as input and returns $C_t' = p(C_t)$, either modifying or leaving it unchanged, rather than moving the reasoning stream forward. Programs are free to execute any logic while performing this task. As such, programs are more general than symbolic tools, though tool usage can be integrated into their execution as desired. For the experiments of this paper, we consider programs that modify $C_t = [C_0, r_{t-1}, s_t]$ through its prompt $C_0$ (e.g.~to add references from a retriever) or its most recent sentence $s_t$ (e.g.~to fix a calculation). See Section \ref{sec:example_programs} for examples.

\textbf{Trigger.} The trigger $\tau$ of a language hook is a text classifier $\tau: C_t \to \{0, 1\}$ that determines whether it would be advantageous to run the hook's program on the existing context. Our framework allows for any text classifier to be used here: an auxiliary LLM, a custom SVM, or even hard-coded rules that detect, for example, mathematical symbols. For the experiments of this paper, our triggers employ the direct-prompting method of \citet{NoisyChannelPrompting} as an approach that generalises well and obviates the need for training hook- or dataset-specific classifiers. Dataset-independent trigger thresholds are selected by analysing a handful of sentences representative of each hook's general use. Selecting thresholds involves a tradeoff between performance and cost; see Appendix \ref{sec:appendix-triggering} for ablations and Appendix \ref{sec:appendix-LH-impl} for example triggers.

\textbf{Eligibility criteria.} The eligibility criteria $e$ of a language hook corresponds to an additional function $e: h_{1...j} \to \{0, 1\}$ that determines whether the hook's program should be permitted to run at the current step in the algorithm. In our experiments, this is a stateful function that allows the hook to run its program at most once between text-generation events and does this through inspecting the history of previously selected hooks $h_{1...j}$ . More generally, it could prevent the program from executing multiple times with similar inputs. Its primary purpose is to avoid pathological behaviour, e.g.~infinite loops, at runtime.

\textbf{Priority.} For the language hook algorithm to be well-defined for a set $H$ of multiple hooks, one more ingredient is required. Since it is possible for more than one hook to be deemed eligible and to trigger at a single step in the algorithm, a priority order $\sigma$ must be defined to break ties in such cases. This could be a user-defined ordering (our approach in this work) or a preference for the hook that triggers with highest confidence, i.e.~highest predicted probability of its text classifier.

\textbf{Algorithm.} Given the ingredients above, the language hook algorithm proceeds event-by-event as follows; see Algorithm \ref{alg:lh}. At each step $j = 1, 2, \ldots$ each hook $h \in H$ is checked for eligibility $e(C_t)$ and for triggering $\tau(C_t)$ on the existing context $C_t$. If multiple hooks satisfy these conditions, then the highest priority one is selected, and its program is executed to modify the context: $C_t' = p(C_t)$. If no hook satisfies these conditions, then the step proceeds as a language-generation event that moves the reasoning stream forward by appending to the context: $C_{t+1} = g(C_t)$. This loop over $j$ continues until a stopping criterion $f(s_t)$ is satisfied. Algorithm \ref{alg:lh} caps the number of iterations at some maximum value $N$, but in practice we set this to an arbitrarily large value and rely on $f$ for termination.

\RestyleAlgo{ruled}


\begin{algorithm}[t]
\caption{Language hook algorithm}\label{alg:lh}\vspace{2mm}
\KwData{Prompt $C_0$, generator $g$, \\ \hspace{1cm}hooks $H = \{h_i = (p_i, t_i, e_i)\}$, \\\hspace{1cm}highest priority hook selector $\sigma$, \\\hspace{1cm}stopping condition $f$.}\vspace{2mm}
\KwResult{Reasoning stream $r_t$.}\vspace{2mm}
$t \gets 1$\;
$C_t \gets g(C_0) = [C_0, s_1]$\;
$r_t \gets \, s_1$\;
 \For{$j \gets 1$ to $N$}{
  \tcp{Get eligible triggered hooks}
  $E_j \gets \{ h_i \in H : e_i(C_t) = 1\}$\; 
  $T_j \gets \{ h_i \in E_j: \tau_i(C_t) = 1\}$\;
  \eIf{$T_j \neq \emptyset$}{
   \tcp{Program execution event}
   $h_j \gets \sigma(T_j)$\;
   $C'_t = p_j(C_t)$\;
   $C_t \gets C'_t$\;
   }{
     \eIf{$f(s_t)$}{
     \tcp{Stopping condition}
     \textbf{break};
   }{
     \tcp{Text generation event}
     $C_{t+1} \gets g(C_t) = [C_t, s_{t+1}]$\;
     $r_{t+1} \gets [r_t, s_{t+1}]$\;
     $t \gets t+1$\;     
     }
   }
 }
\Return $r_t$  
\end{algorithm}

\subsection{Examples of language hook programs}
\label{sec:example_programs}

Here we describe the program for each language hook used in the experiments of Section \ref{experiments}. Further details, including auxiliary-LLM prompts, are provided in Appendices \ref{sec:appendix-LH-impl} and \ref{sec:appendix-hook-prompts-program}.

\label{calculator}
\textbf{Calculator hook.} LLMs can still struggle  with arithmetic tasks \cite{MATH}, particularly with large numbers  \cite{LimitsofLMsforSimpleArithmetic, LimitsLMsArithmeticSymbolicInduction}. Our calculator program aims to fix any incorrect calculations in the most recent sentence $s_t$. The program first prompts an auxiliary LLM to extract all calculations from the sentence; these are then verified using the symbolic tool SymPy \cite{sympy}. If an error is detected, the program corrects it and removes further reasoning to prevent errors propagating to future text generation. 
If no errors are detected, the program leaves the context unchanged.

\label{retriever} 
\textbf{Retriever hook.} The retriever program searches for relevant references to add to the prompt $C_0$ and rewrites the most recent sentence $s_t$ in light of the modified context. More specifically, the program:
(i) prompts an auxiliary LLM to generate 5 search queries based on $C_{t-1}$; 
(ii) feeds these queries to a retriever tool; 
(iii) prompts the auxiliary model to propose a rewrite of $s_t$ based on the retriever output;
(iv) if the rewrite cites a reference, then update $s_t$ accordingly and add the citation to $C_0$; otherwise leave the original context $C_t$ unchanged.

\label{guardrail}
\textbf{Guardrail hook.} At present, LLMs like ChatGPT \cite{chatgpt} are desirable for use as base models in text-generation applications, but they are often reluctant to engage with a prompt if there is not enough information to respond accurately. Such internal validation is necessary for models that have no access to tools, but it can be cumbersome and unnecessarily limiting when the base model is augmented with external methods of validation instead, as is the case with our method. In fact, the language hook framework offers a transparent way to intercept and redirect occurrences of the base model's ``guardrails''. Upon triggering, our guardrail program simply prompts the base model to make its best guess at reasoning so that the reasoning stream can continue; see Appendix \ref{sec:appendix-LH-impl} for further details. Variants of this simple hook offer a way to externalise validation of the base model; this will become increasingly appropriate as LLMs become more widely integrated with external tools.

\section{Experiments} \label{experiments}

In this section, we validate the language hooks framework by benchmarking the specific hooks introduced in Section \ref{sec:example_programs} against state-of-the-art general and task-aware approaches in three different domains: mathematical reasoning, multi-hop QA, and a custom composite dataset that requires a combination of tools for good performance.

\subsection{Experimental setup} \label{experimental-setup}

\input{table_reasoning}

We implement the language hook algorithm with ChatGPT (\texttt{gpt-3.5-turbo-0301}) \cite{chatgpt} as the base model, \texttt{curie} for triggering, and ChatGPT as the auxiliary LLM within programs. We place our hooks in decreasing priority order as: retriever, guardrail, calculator. See Appendix \ref{sec:appendix-LH-impl} for the implementation of each hook in detail.

We compare the language hooks method to two general techniques: Chain-of-Thought Prompting (CoT) \cite{CoT} which has no access to tools, and ReAct \cite{ReAct} which uses the prompt to demonstrate how the base model may initiate tool usage by outputting specific token patterns. We also compare against two strong task-aware baselines, PAL \cite{PAL} for mathematical reasoning and DSP \cite{DSP} for multi-hop QA, which we comment on further in the subsections below.

Unless noted otherwise, in our experiments we ensure that all methods employ the same base model, the same in-context examples, and the same specific tools. For each method, the number of in-context examples (randomly selected from the training split of each dataset) is $\min{(3, k)}$ where $k$ is the maximum number of examples the method can support within context limits.\footnote{This primarily affects ReAct, which has room for 2 in-context examples in Section \ref{multi-hop} and 1 in Section \ref{composite-QA}.} Full prompts for all methods are included in Appendix \ref{sec:appendix-hook-prompts-base}.

For each benchmarking experiment, we report performance on 500 unseen questions to control costs; these examples are randomly sub-sampled either from the dataset's official test set (when one exists) or from the validation split.

\subsection{Mathematical reasoning} 
\label{math-reasoning}

In this first set of experiments, we benchmark the calculator hook introduced in Section \ref{calculator}.

\textbf{Baselines.} Alongside CoT and ReAct we also compare against PAL \cite{PAL}. PAL prompts the base model to write a computer program to solve each question, and the program is fed to a Python interpreter. We use the official implementation of PAL and the \texttt{gpt-3.5-turbo-0301} support provided in their codebase. 

\textbf{Tool.}
Aside from PAL (which executes Python) both language hooks and ReAct are paired with SymPy. Given that all questions require only basic arithmetic this difference solely relates to the interface, not the capability, of the tool. 

\textbf{Datasets.} 
We evaluate on the GSM8K dataset \cite{GSM8K}, containing grade school math word problems, and on GSM-HARD \cite{PAL}, where numbers in GSM8K are artificially made larger to make the calculations more challenging. However, this creates many questions with nonsensical reasoning paths (e.g.~negative salaries) which \texttt{gpt-3.5-turbo-0301} often refuses to answer (see Appendix \ref{sec:appendix-gsmhard_questions} for examples). We therefore filter the 500 questions in our GSM-HARD evaluation set to remove cases where the answer is negative or non-integer, leaving 326 questions that we refer to as GSM-HARD-filtered. We report results in Table \ref{table:math-results}. 

\textit{Language hooks are non-intrusive.} On GSM8K, where a calculator isn't strictly needed given \texttt{gpt-3.5-turbo-0301}'s ability to do simple arithmetic, the running of the calculator program to validate the calculations in the generated reasoning stream does not harm performance -- i.e.~the hook performs validation in a fault-free way. In contrast, Table \ref{table:math-results} shows that ReAct degrades performance compared to CoT, which we hypothesise is due to ReAct's strict response structure which differs sharply from the model's pre-training data.

\textit{Language hooks are effective.} On GSM-HARD language hooks performs slightly behind PAL, although this is not surprising as PAL, by translating the problem into code, abstracts away numerical values making it less sensitive to the implausible reasoning steps required on GSM-HARD. 
Once we adjust for this in GSM-HARD-filtered, language hooks in fact performs slightly ahead of PAL.

\subsection{Multi-hop QA} \label{multi-hop}

\input{table_multihop}

In these experiments, we benchmark the guardrail and retriever hooks described in Section \ref{retriever}.

\textbf{Baselines.} 
As above, we include the general methods CoT and ReAct. In this experiment, we also allow ReAct to fall back on CoT if it exceeds the context-length limit before satisfying the stopping criterion (i.e.~returning a final answer).\footnote{This matches the method laid out in \citet{ReAct} aside from the fact that, to control costs and maintain comparability with language hooks and CoT itself, we did not allow ReAct to fall back on CoT with self-consistency \cite{self-consistency}.}

We also benchmark DSP \cite{DSP}, an approach that creates sophisticated pipelines between the base LLM and a retriever when solving knowledge-intensive QA tasks. We use the official implementation with \texttt{gpt-3.5-turbo-0301} support, keeping all other settings at their defaults (apart from the external retriever, which we align across all the methods tested). As a result, DSP is the only method in our experiments that benefits from self-consistency \cite{self-consistency}. It is also the only method that controls the creation of its own in-context examples, as it has distinct requirements here that form a core part of the method.

\textbf{Tool.} We employ the hybrid retriever from the \texttt{retriv} package with default hyperparameters \cite{retriv} and build an index for each dataset using the titles and passages combined. For language hooks and ReAct, a single retriever query returns the 3 closest results, and for DSP we allow this to vary based on their implementation. To alleviate issues associated with large retrieved documents, we truncate each to the first 128 words.

\textbf{Datasets.}
We evaluate on two challenging multi-hop QA datasets, HotpotQA \cite{HotpotQA} and 2WikiMultihopQA \cite{2WikiMultihopQA}. For the former, we evaluate in the open-domain multi-hop (``fullwiki'') setting where a retriever must search across the whole of Wikipedia to obtain supporting facts; in particular, we use the official 2017 supporting facts (5,233,329 unique documents) released with the dataset. 2WikiMultihopQA is originally a reading comprehension task which, following \citet{InterleavingRetrievalWithChainOfThought}, we turn into a more challenging open-domain task, combining all supporting passages from the train, development, and test sets which the retriever must search against (398,354 unique documents). See Table \ref{table:multihop-results} for results.

\textit{Language hooks outperform task-aware methods.} Table \ref{table:multihop-results} shows that language hooks achieves comparable, often significantly improved, performance to the state-of-the-art prompt-based approaches tested, including DSP which is a strong task-aware baseline using self-consistency. 

\textit{Language hooks are modular.} Language hooks also outperform ReAct. Due to their modularity, they alleviate the ``distractibility'' \cite{LargeLanguageModelsDistractedByIrreleventContext} that ReAct suffers from due to its requirement that the base model's prompt/context must contain the task description, tool-use instructions, and even tool (e.g.~search query) outputs; see Listing \ref{lst:hotpotqa_react_prompt} for an example prompt. We also find (as noted by the authors themselves) that around half of ReAct's failures stem from the method getting stuck in repetitive loops. In contrast to these issues, language hooks allows the base model to focus on the primary task at hand as described in its tool-agnostic prompt. Nuanced instructions and tool-specific examples are provided to an auxiliary LLM within the retriever hook, allowing better performance at the narrow retrieval task. This is accomplished without distracting the base model and taking up minimal space in its context.

\begin{figure}
    \centering
    \includegraphics[width=0.9\linewidth]{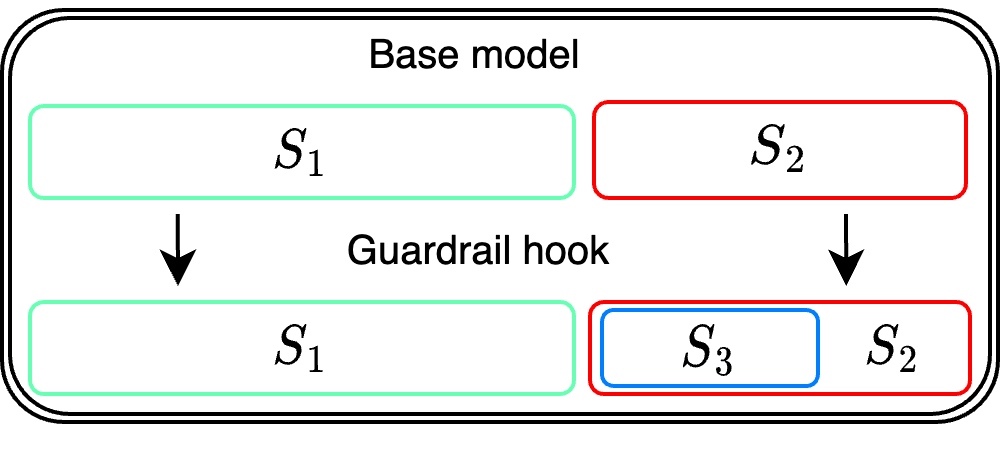}
    \caption{Setup for results in Table \ref{table:ablation-guardrail}. We run the base model with no active hooks to identify $S_1$, those questions which the base model answers, and $S_2$, those which it refuses to answer. Our guardrail hook then defines a subset $S_3 \subseteq S_2$, which the base model now answers but previously did not.}
    \label{fig:guardrail-ablation}
\end{figure}

\textit{Language hooks offer external validation.} Here we analyse the potential for language hooks to augment internal safety measures in the base model (\texttt{gpt-3.5-turbo-0301}). From the HotpotQA and 2WikiMultihopQA evaluation sets, we consider three subsets of questions: $S_1$, those which the base model  answers; $S_2$, those which the base model refuses to answer; and $S_3 \subseteq S_2$, those which our guardrail hook intercepts and redirects so that the base model then proceeds with an attempted response. Figure \ref{fig:guardrail-ablation} depicts this setup, and Table \ref{table:ablation-guardrail} displays the performance of different combinations (or not) of language hooks on $S_1$ and $S_3$ (whereas $S_2$ contains unanswered questions).

\input{./table_guardrail}

Note that, without an active retriever hook, the base model performs much more strongly on $S_1$ than $S_3$. As $S_3$ represents the subset of questions that the base model initially refused to answer but was coerced to respond to anyway, this shows the importance and effectiveness of \texttt{gpt-3.5-turbo-0301}'s internal validation capacity. On the other hand, note the dramatic improvement on $S_3$ that results from incorporating our retriever hook, which leads to F1 scores comparable to -- or even exceeding -- the base model's performance on $S_1$. This is performance that would be left on the table without an active hook to intercept the base model's guardrails. This result shows the promise of coupling a less constrained base model with language hooks for external validation of its outputs; see Section \ref{sec:ethics} for further discussion.

\subsection{Composite QA} \label{composite-QA}

In these experiments, we benchmark the guardrail, calculator, and retriever hooks in combination. We compare against CoT and ReAct, the latter of which uses the calculator and retriever tools as discussed in Sections \ref{math-reasoning} and \ref{multi-hop}.

\textbf{Dataset.} The ability of a language model to synthesise responses to individual sub-questions into a final answer is under active research \cite{SelfAsk, faith-and-fate}. We create a custom task, HotpotQA$\times$GSM8K, to investigate how well different approaches utilise multiple tools in this setting. In particular, we select questions from HotpotQA and GSM8K where the answer is an integer value with more than 3 significant figures; we then pair these questions, one from each dataset, to create composite questions where the final answer requested is the {product} of the answers to the individual questions (see Appendix \ref{sec:appendix-composite_questions} for examples). Answering a composite question thus requires interleaving mathematical reasoning with multi-hop QA along with a nontrivial large-integer multiplication step at the end. Good performance on this task requires an understanding of {when} to use {which} tool and how to properly compose the final answer. The task thus evaluates each method's ability to combine multiple tools in a novel and more realistic setting. See Table \ref{table:multihook-results} for results.

\input{table_multihook}

\textit{Language hooks can combine tools effectively in unseen settings.} The results in Table \ref{table:multihook-results} show that the language hook framework adapts to multiple-tool use with no explicit demonstration of such, significantly outperforming ReAct in this setting. We can further analyse this finding through an ablation that assesses the performance of ReAct and language hooks on the individual components of each composite question, in the case that each method is only given access to the single (correct) tool for each sub-question. The results of this ablation are provided in Table \ref{table:ablation-composite-simple}. 

\input{./table_ablation_composite_simple}

If a method is able to both select the correct tool for each sub-task and compose the final answer from its multi-part response, then the expected composite performance in Table \ref{table:ablation-composite-simple} will be close to the performance in Table \ref{table:multihook-results}. We find a relative 31.1\% drop in performance for ReAct and an 18.8\% drop for language hooks. Further ablations are provided in Appendix \ref{sec:appendix-analysis} (see Tables \ref{table:ablation-composite-full} and \ref{table:ablation-composite-deconstructed}) where we isolate a more dramatic difference in the two methods' abilities to compose their final answers.

The results of this section validate that the language hook framework adapts to novel composite domains significantly better than other approaches. 

\section{Related work}

In this section we discuss the large body of recent literature relating to our work.

\textbf{Task-aware methods for tool usage.} Many works have created task-aware control-flows to solve narrow problems with the high-level steps required specified in code \cite{InterleavingRetrievalWithChainOfThought, DSP, SearchInTheChain}. Another approach utilises the prompt to demonstrate how to decompose a problem and/or use tools. This appears in a variety of forms, from a single model call \cite{PAL, SolvingMathWordProblems}, to chained model calls \cite{DecomposedPrompting, ReAct,  ART, Chameleon, SearchInTheChain, ChatCoT}, and repeated prompting \cite{LeastToMostPrompting, MaieuticPrompting, SuccessivePrompting}. All these approaches rely either on externalising the logical flow into code or on providing explicit examples of tool use on specific tasks, limiting generalisability in either case. In contrast, language hooks flexibly adapt the control-flow at runtime, based on the evolving context and the available programs, and do not rely on tool demonstrations in the base prompt, which are liable to inadvertently steer the base model towards undesired reasoning paths. Our method is task-agnostic and non-intrusive in this way.

\textbf{Model-specific methods for tool use.}
Language models can also learn to use tools through fine-tuning \cite{TALM, Toolformer}, and this does allow the model to fully dictate the control-flow. Related works such as ToolkenGPT \cite{ToolkenGPT} learn tool embeddings or use distillation approaches to concentrate a model's abilities on narrower tasks \cite{SpecializingSmallerLanguageModels}. The downsides of all these approaches is of course that tool competency gets coupled to the fine-tuned model; this makes upgrading the base model, and even rotating tools, an expensive event. In contrast, language hooks are model-agnostic, requiring no additional training data and involving no parameter updates to the base model.

\textbf{Single-tool specialised methods.}
Several recent methods \cite{Retrieval-Augmented-LMs, LeveragingPassageRetrieval, Few-shot-Retriever, REPLUG} offer ways to equip a language model with up-to-date knowledge through the use of a retriever, rather than relying on the parametric memory of the base model \cite{Internet-augmented-LMs, RealTimeQA}. In particular, FLARE \cite{FLARE} iteratively generates a candidate sentence and optionally performs retrieval using it; SearChain \cite{SearchInTheChain} can additionally correct invalid reasoning steps; SeeKeR \cite{LanguageModelsThatSeekKnowledge} chains many LLM calls together to generate a response; and IRCoT \cite{InterleavingRetrievalWithChainOfThought} bears resemblance to our retriever program in the way that it prepends retrieved information to the generated text. However, all of these approaches are retriever-focused and thus narrower in scope than language hooks. Our method provides a modular framework for augmenting the base model with any number of new capabilities.

\textbf{Other contemporaneous work.} Similar to our calculator program, \citet{SolvingMathWordProblems} uses an LLM to formalise word problems and pass the output to an external symbolic solver. This is another single-tool method distinct from our modular approach. Also concurrent to our work, \citet{CRITIC_self_correction} introduce CRITIC, which allows a language model to iteratively refine its initial response. By contrast, language hooks identify and correct mistakes as they occur without allowing errors to propagate to future text generation. MoPE \cite{MoPE} ensembles multiple specialized LLMs using a trained classifier to select the best response. Importantly, language hooks flexibly modulate between different augmented capabilities without requiring any particular one to be sufficient for generating a full response.

\section{Conclusion}

In this paper we have introduced language hooks, a flexible and generic framework for equipping language models with new capabilities. The idea behind our method is natural: stream text out of a language model and use external programs to validate and augment the generated text in tandem. However, the implications are far-reaching, including the decoupling of tool usage from both the base model and its prompt. Beyond this, for certain applications we expect external validation of language model outputs to become the dominant paradigm; language hooks offer a transparent and modular way to accomplish this.

Our experiments show that language hooks achieve performance competitive with both general and task-aware benchmarks, comparing favourably to strong baselines. We validated this on both mathematical-reasoning and multi-hop QA tasks. We also demonstrated language hooks' generalisability, outperforming baselines on a novel task requiring multi-tool use.

Avenues for future work include further extending the functionality of language hooks beyond tool usage. This could involve hooks that detect and redirect toxic language or age-restricted content. One could also take advantage of the conditional computation that language hooks enable in a variety of different ways. For example, a hook could be developed to apply self-consistency in a more localised and cost-efficient fashion. In general, we hope our work leads to further advancements that enable augmented language models to have a positive impact across a wide range of applications.

\clearpage
\section{Ethical considerations} \label{sec:ethics}

\textbf{External validation.} Language hooks offers a framework for the external validation of model outputs, independent from the base model and in a transparent manner, at the point of text generation. We believe this has compelling applications, and the language hooks framework could be used in conjunction with internal safety measures (e.g.~data collection and RLHF \cite{rlhf}) as a viable route for external validation of model outputs. Indeed, we believe safety is a property of the system as a whole, not solely of the language model, and that language hooks provides a framework to support the governance of language models, particularly for safety critical applications. 

\textbf{Guardrail hook. }The guardrail hook prompted the base model to respond when it has initially determined it did not have the required information. Prompting a model to guess in this way introduces the risk that the model will hallucinate information and therefore, as done in our experiments (see Appendix \ref{sec:appendix-guardrail}), it can be beneficial to mark this text as potentially untrustworthy if used in later text generation steps. 
More generally, we believe that language models should not be solely relied upon as a source of information, and we encourage further research efforts into methods of transparently connecting them with external tools to allow for independent and transparent verification of their outputs.

\section{Limitations}

\textbf{Interaction of multiple hooks.} Our experiments demonstrate the use of multiple tools when solving a single task, which requires the setting of a priority order between hooks. A key consideration when designing programs is to avoid destructive interference where the program from one hook may undo or contradict the modifications made by another hook's program.

\textbf{Usage via. API.}  The current OpenAI API does not implement any mechanism for preserving the hidden state from model calls. As such, upon triggering, our experiments require re-calculating the context for every sentence, thus increasing the cost of using language hooks under the current OpenAI API.

\textbf{Greedy approach to generation.} The language hook algorithm generates text one sentence at a time. This is essentially a greedy text generation approach which doesn't allow for the LLM to plan for tool usage several steps ahead of time. We imagine that the language hook framework may be ill-suited for some class of tools, where awareness of the availability of the tool is beneficial.

\bibliography{custom}
\bibliographystyle{acl_natbib}

\clearpage
\appendix
\section{Further analysis and ablations} \label{sec:appendix-analysis}
\input{./table_ablation_composite}
\input{./table_ablation_composite_deconstructed}

\subsection{Tool selection} \label{sec:appendix-tool-selection}
We run the following ablations for ReAct and language hooks on the individual sub-questions in the composite QA dataset from Section \ref{composite-QA} to assess their performance: \textbf{(A)} when given access to the correct tool only (as in Table \ref{table:ablation-composite-simple}) and \textbf{(B)} when given access to all tools (multi-tool).

Ablation \textbf{(A)} gives the baseline performance of an approach for each individual sub-question. Ablation \textbf{(B)} assesses an approach's ability to know when to use the correct tool, or ignore redundant ones. The results are in Table \ref{table:ablation-composite-full}.

We see that both ReAct and language hooks perform similarly under \textbf{(B)} as with \textbf{(A)}, demonstrating both approaches are able to correctly select tools when given a choice. As such the main performance drop comes from their ability to compose the individual answers together, which we analyse in the next section. 

\subsection{Compositionality} \label{sec:appendix-compositionality}
Whilst the above ablations allowed us to test tool selection, we can analyse how well approaches can compose their responses by analysing the results from the composite QA dataset which were presented in Table \ref{table:multihook-results}. To recap: a composite question consists of a GSM8K question's answer multiplied by a HotpotQA question's answer. Therefore to correctly answer a composite question requires: 

\begin{enumerate}[noitemsep]
    \item Answering both sub-questions correctly;
    \item Extracting the above answers from the reasoning to input into the final multiplication;
    \item Correctly multiplying the answers of the sub-questions together to form the final answer.
\end{enumerate}

We classify steps 2 and 3 above as the \textit{compositionality gap} \cite{SelfAsk}. Step 2 is subtle but necessary since it is possible, for example, for a model to answer a sub-question correctly but then to retrieve an incorrect value for the final multiplication. 

The results are shown in Table \ref{table:ablation-composite-deconstructed}. We see that CoT correctly answers both sub-questions 36.7\% of the time whereas 35.3\% of the time it achieves this as well as retrieves the answers to the sub-questions to input into the final multiplication (a drop of 1.4\%). However, out of the questions in which step 2 has been correctly completed, CoT only correctly completes the multiplication of these numbers 53.8\% of the time. By contrast, ReAct drops 8\% in performance (39\% - 31\%) from its inability to retrieve previously correct answers from its reasoning per step 2 above. We hypothesise this is a type of self-distraction which ReAct suffers from more than language hooks (which only drops 2.7\% at step 2), given that it must attend over a large context, containing initial prompts with tool demonstrations, retrieved information, reasoning and answers to sub-questions when composing its final multiplication.

\subsection{Trigger analysis} \label{sec:appendix-triggering}

\textbf{Performance sensitivity to triggering threshold. }
We run ablations to test the sensitivity to the trigger threshold used in our experiments. We do this for the mathematical reasoning datasets from Section \ref{math-reasoning} and the multi-hop QA datasets from Section \ref{multi-hop}. In our experiments, the calculator hook triggers on around 70\% of sentences in the mathematical reasoning datasets and the retriever hook on around 95\% of the sentences in the multi-hop QA datasets\footnote{This is acceptable as almost every sentence requires external information on the multi-hop QA datasets.}. As such, we choose three new thresholds:
(i) \textbf{very low rate}: both hooks trigger around 20-30\% of the time; 
(ii) \textbf{low rate}: both hooks trigger around 50\% of the time;
(iii) \textbf{high rate}: calculator hook triggers 90-95\% of the time, and retriever hook triggers > 99\% of the time;

\begin{figure*}[t]
  \centering
\includegraphics[width=\textwidth]{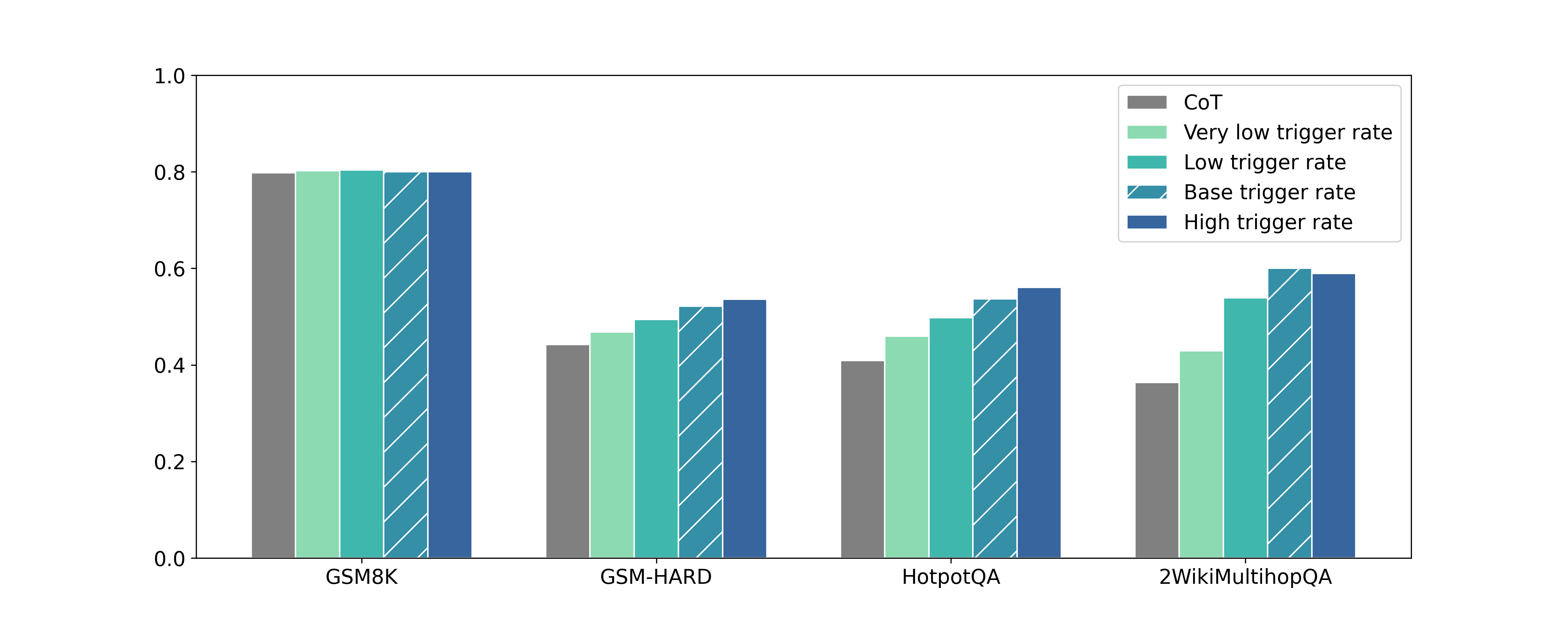}
\caption{F1 score per dataset as we vary the trigger threshold for language hooks. CoT and base trigger rates correspond to results quoted in Tables \ref{table:math-results} and \ref{table:multihop-results}.}
  \label{fig:trigger}
\end{figure*}

The results are shown in Figure \ref{fig:trigger} where we see that decreasing the rate at which a hook triggers drops performance towards the level of CoT as expected. Increasing the trigger rate activates the hooks more frequently and improves performance. Further, over-triggering does not degrade performance, demonstrating programs are robust to this. This robustness means one has an element of control in choosing thresholds and can err on the side of being trigger-happy if one’s budget allows for it.

\begin{figure*}[t]
  \centering
\includegraphics[width=\textwidth]{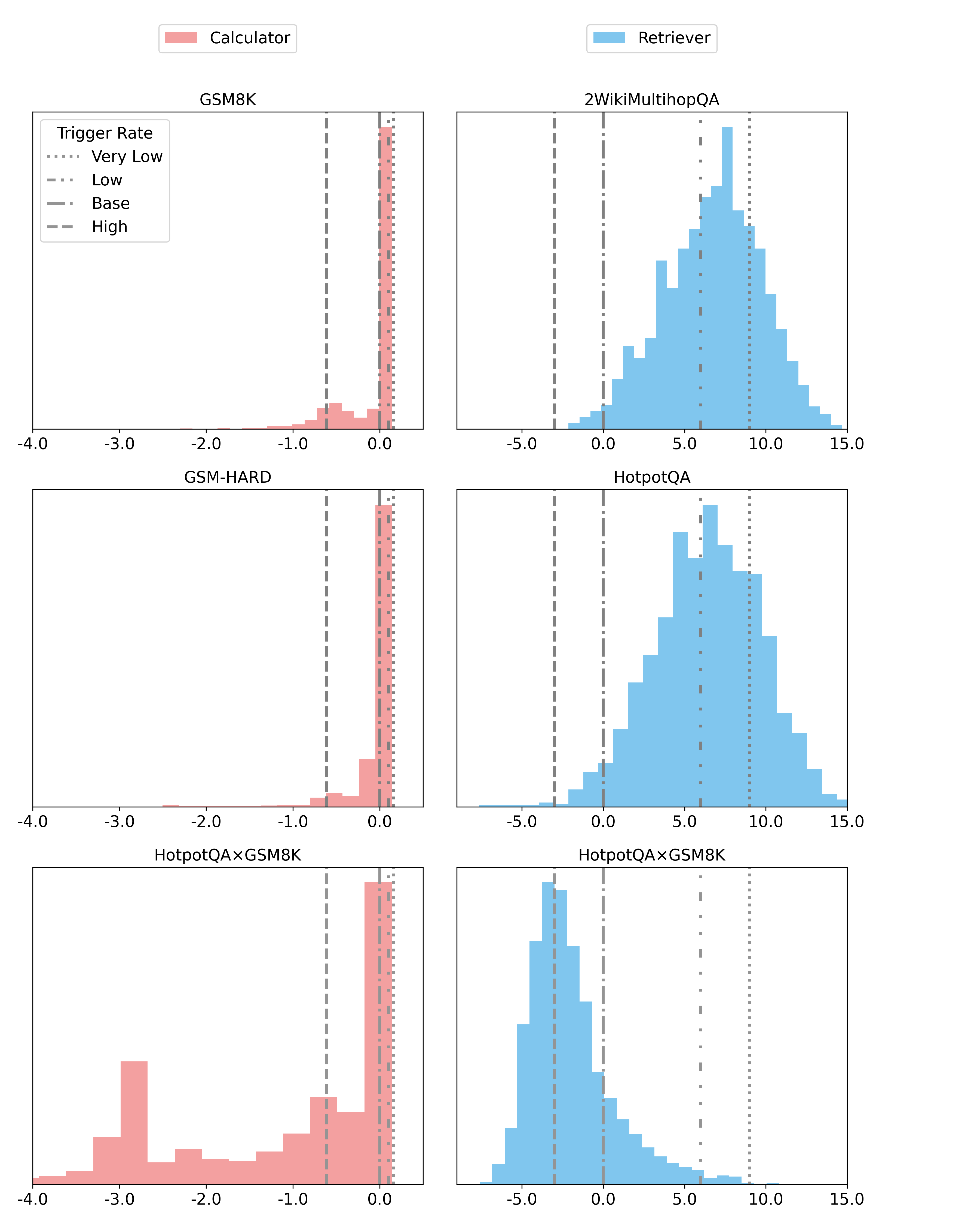}
\caption{\textbf{Trigger probability distributions}. The left and right hand columns show the probability distributions for the calculator trigger (light red) and the retriever trigger (blue) respectively. A hook runs its program when $P(\text{trigger}) > \text{threshold}$. We show the base trigger threshold used in our main experiments from Section \ref{experiments} at 0.}
  \label{fig:trigger_probs}
\end{figure*}

\textbf{Trigger probability distributions. } Figure \ref{fig:trigger_probs} shows the triggering probability distributions for the calculator and retriever hooks on the different datasets. We can see that whether to trigger the calculator hook is essentially a binary decision (hence a bimodal distribution on the composite dataset) since it's simply validating whether a sentence contains mathematical calculations or not. On the other hand, determining whether to trigger the retriever hook, which must determine if extra information is required, is a continuum for which we must choose a cut-off. 

\clearpage

\clearpage
\section{Language hooks: implementation details} \label{sec:appendix-LH-impl}
In this section we provide further details around the implementation of the language hooks used in our experiments in Section \ref{experiments}.

\subsection{Calculator hook} \label{sec:appendix-calculator}

\subsubsection{Program}
The calculator program fixes any incorrect calculations in the most recent sentence of the context $s_t$. This is accomplished through the following steps:
\begin{enumerate}[noitemsep]
  \item \textbf{Extract} any calculations in the last sentence. This is accomplished by prompting \texttt{gpt-3.5-turbo-0301} with the prompt shown in Listing \ref{lst:program_calculator_extract}.
  \item \textbf{Format} extracted calculations to be parsed by Python. This is accomplished by prompting \texttt{gpt-3.5-turbo-0301} with the prompt shown in Listing \ref{lst:program_calculator_format}.
  \item \textbf{Validate} calculations using the SymPy Python library.
  \item \textbf{Correct} calculations in the original sentence. This step only occurs if the validate step identified wrong calculations. This is accomplished by prompting \texttt{gpt-3.5-turbo-0301} with the prompt shown in Listing \ref{lst:program_calculator_correct}.
  \item \textbf{Trim} text after the calculation so as to avoid propagating mathematical errors. This is accomplished by prompting \texttt{gpt-3.5-turbo-0301} with the prompt shown in Listing \ref{lst:program_trim}.
\end{enumerate}
\subsubsection{Trigger}
The trigger for the calculator hook classifies between sentences containing calculations and sentences not containing calculations. This is accomplished by evaluating the OpenAI \texttt{curie} model using the prompt shown in Listing \ref{lst:calculator_trigger}. Sentences for which the verbaliser has a model log-probability above -0.14 are triggered on.
\subsection{Retriever hook} \label{sec:appendix-retriever}
\subsubsection{Program}
The retriever program enables the language model to rewrite sentences using knowledge found in external knowledge bases. This is accomplished through the following steps:
\begin{enumerate}[noitemsep]
  \item \textbf{Generate} a set of (5) relevant search queries which can be used to find external references. This is accomplished by prompting \texttt{gpt-3.5-turbo-0301} with the prompt shown in Listing \ref{lst:program_generate_search}.  
  \item \textbf{Search} using a retriever connected to a knowledge base. In our experiments the top 3 references are kept for every search query and duplicate references are removed. This leads to up to 15 references.
  \item \textbf{Re-generate} the last sentence with all references added to the context. For this step the prompt is constructed in a way that encourages the language model to add explicit citations to references. This is accomplished by prompting \texttt{gpt-3.5-turbo-0301} with the prompt shown in Listing \ref{lst:program_regenerate_last_sentence}.  
  \item \textbf{Check} whether the generated sentence contains citations to references. If yes, overwrite the original sentence, otherwise fallback on the original sentence.
  \item \textbf{Augment} the context, $C_0$, with any newly added references. Here we only keep references actually cited and remove all other references.
\end{enumerate}
\subsubsection{Trigger}
The trigger for the retriever hook trigger classifies whether the last sentence contains information that could have been found using an external knowledge source (e.g.~Wikipedia). This is accomplished by prompting the OpenAI \texttt{curie} model with the prompt shown in Listing \ref{lst:retriever_trigger}. Sentences where the verbaliser has a log-probability above -25 are triggered on.

\subsection{Guardrail hook} \label{sec:appendix-guardrail}
\subsubsection{Program}
\texttt{gpt-3.5-turbo-0301} is often unwilling to answer questions if it believes that its answers may contain incorrect knowledge or the question isn't well-posed. This behaviour becomes exacerbated when generating the text sentence-by-sentence, as required in the language hook framework. The guardrail hook's purpose is to nudge the language model to attempt to answer all questions, even those it is unsure of whilst making it clear in the generated text where such guesses may occur. Concretely we do this through the following steps:
\begin{enumerate}[noitemsep]
  \item \textbf{Augment} the context with a partially completed sentence "From reference [x] we learn that" where [x] is a fictitious reference not given to the model.
  \item \textbf{Complete} the sentence using the LLM (\texttt{gpt-3.5-turbo-0301}). If, for this step, the context $C_0$ already contains references we temporarily suppress them.
  \item \textbf{Replace} "From reference [x] we learn that" in the newly generated sentence with "We make a guess that". This step ensures that later sampling steps are aware that the knowledge elicited by the guardrail hook is potentially untrustworthy. 
\end{enumerate}
\subsubsection{Trigger}
The goal of the guardrail hook trigger is to classify whether the language model has refused to answer the question in the last sentence. This is accomplished by prompting the OpenAI \texttt{curie} model with the prompt shown in Listing \ref{lst:guardrail_trigger}. Sentences where the verbaliser has a log-probability above -0.5 are triggered on.

\subsection{Eligibility criteria} \label{sec:appendix-elibility}

All of the hooks introduced in the paper were allowed to run their program only once between text generation events. This eligibility criteria ensures that any generated text is validated by each hook at most once and avoids any risk of infinite loops and wasted computation.

\subsection{Priority order} \label{sec:appendix-priority}

In all of our experiments we assigned the highest priority to the retriever hook, the next highest priority to the guardrail hook and the lowest priority to the calculator hook. We set the retriever hook to have higher priority than the guardrail hook to allow the model to benefit from external information before being prompted to make a guess by the guardrail hook. The calculator hook is given the lowest priority, thus only validating calculations when the factual content of the sentence has been determined.

\subsection{Stopping condition} \label{sec:appendix-stopping}

The language hook algorithm continues to run until either the pre-specified maximum number of event calls is reached or a user-defined stopping criterion is satisfied. The stopping criterion is only evaluated if there are no eligible triggered hooks.

The stopping criterion used in our experiments does this by checking $s_t$ for either an end-of-sequence token or an answer (using string matching with the answer structure found in the in-context examples). If any of the above conditions are satisfied then the algorithm exits.

\subsection{Sentence splitting} \label{sec:appendix-sentence_splitting}

We make use of the NLTK toolkit \cite{nltk_codebase} which implements the sentence splitting algorithm introduced in \citet{sentence_boundary_detection_algorithm}.

\clearpage
\onecolumn
\colorlet{shadecolor}{gray!10}
\lstset{breaklines=true, columns=fullflexible, backgroundcolor=\color{shadecolor},  keywordstyle=\color{black},  stringstyle=\color{black}, inputencoding=utf8/latin1}

\section{Dataset (example questions)} \label{sec:appendix-example-questions}

\subsection{GSM-HARD} \label{sec:appendix-gsmhard_questions}

\begin{lstlisting}[basicstyle=\footnotesize, caption={Example question from GSM-HARD}, label={lst:gsm_hard_1}]
Hannah has a mental breakdown while studying for finals and starts smashing windows in the high school's parking lot. She smashes a quarter of the students' cars' windows and 3/4ths of the teachers' cars' windows. If there are 3977391 students' cars with four windows each and 32 teachers' cars with two windows each, how many windows does Hannah smash?
\end{lstlisting}
\begin{lstlisting}[basicstyle=\footnotesize, caption={Example question from GSM-HARD}, label={lst:gsm_hard_2}]
Marcus ordered 5 croissants at $7808570.00 apiece, 4 cinnamon rolls at $2.50 each, 7808570 mini quiches for $4.00 apiece and 17808570 blueberry muffins that were $1.00 apiece.  At check out, Marcus shows his loyalty card that gives him 10% off of his purchase.  What is Marcus' total bill?
\end{lstlisting}
\begin{lstlisting}[basicstyle=\footnotesize, caption={Example question from GSM-HARD}, label={lst:gsm_hard_3}]
Jayden had $2678923 from selling pictures he took as a hobby.  His sister Ava gave him half of her $90 allowance to help him buy a new camera that costs $200. How much more does Jayden need to buy the camera?
\end{lstlisting}

\subsection{Composite QA} \label{sec:appendix-composite_questions}
\begin{lstlisting}[basicstyle=\footnotesize, caption={Example question from composite QA dataset}, label={lst:composite_1}]
What is the answer to the first part multiplied by the answer to the second part?
First part: Grace weighs 125 pounds. Alex weighs 2 pounds less than 4 times what Grace weighs. What are their combined weights in pounds?
Second part: What was the largest passenger capacity of the plane type used for BOAC Flight 911 ?
\end{lstlisting}
\begin{lstlisting}[basicstyle=\footnotesize, caption={Example question from composite QA dataset}, label={lst:composite_2}]
What is the answer to the first part multiplied by the answer to the second part?
First part: Two trains leave San Rafael at the same time. They begin traveling westward, both traveling for 80 miles. The next day, they travel northwards, covering 150 miles. What's the distance covered by each train in the two days?
Second part: Brown State Fishing Lake is in a country that has a population of how many inhabitants ?
\end{lstlisting}
\begin{lstlisting}[basicstyle=\footnotesize, caption={Example question from composite QA dataset}, label={lst:composite_3}]
What is the answer to the first part multiplied by the answer to the second part?
First part: What was the Roud Folk Song Index of the nursery rhyme inspiring What Are Little Girls Made Of?
Second part: Meredith is a freelance blogger who writes about health topics and submits to clients each day as her permanent job. A blog article takes an average of 4 hours to research and write about. Last week, she wrote 5 articles on Monday and  2/5 times more articles on Tuesday than on Monday. On Wednesday, she wrote twice the number of articles she wrote on Tuesday. Calculate the total number of hours she spent writing articles in the three days.
\end{lstlisting}

\section{Prompts: trigger} \label{sec:appendix-hook-prompts-trigger}

The below Listings show the prompt templates for the trigger classifiers used in our experiments. Text in \textcolor{red}{red} denotes values which are placeholders for question-specific content either from the question itself or the rationale generated by the model. Text in \textcolor{blue}{blue} denotes verbaliser tokens over which trigger probabilities are calculated. Text in black is part of the prompt template. 

\begin{lstlisting}[basicstyle=\footnotesize, , escapeinside={(*@}{@*)}, caption={\textit{Calculator hook.} Verbaliser tokens over which trigger is evaluated are shown in \textcolor{blue}{blue}. Placeholders for question-specific content are shown in \textcolor{red}{red}.}, label={lst:calculator_trigger}]
Classify each sentence as containing or not containing mathematical calculations.
The sentence "Firstly let's try to answer the first question step by step." contains no mathematics
The sentence "Bob already had 5 books and buys 10 books which means he now owns 10+5 = 15 books." contains mathematics
The sentence "According to reference [1] the population of the USA in 2005 was 2000000." contains no mathematics
The sentence "Mario's original salary was 4000/1.4=2857.14 dollars" contains mathematics
The sentence "So she needs to copy 20*2=40 pages before the end of the day." contains mathematics
The sentence "The answer to the first question is 2005." contains no mathematics
The sentence "The second world war started on the 18 of March 1939." contains no mathematics
The sentence "(*@\textcolor{red}{\{last sentence\}}@*)" contains (*@\textcolor{blue}{ mathematics}@*)
\end{lstlisting}

\begin{lstlisting}[basicstyle=\footnotesize, escapeinside={(*@}{@*)},  caption={\textit{Retriever hook.} Verbaliser tokens over which trigger is evaluated are shown in \textcolor{blue}{blue}. Placeholders for question-specific content are shown in \textcolor{red}{red}.}, label={lst:retriever_trigger}]
Question: (*@\textcolor{red}{\{question\}}@*)
Answer:
Let's think step by step. (*@\textcolor{red}{\{rationale\}}@*) In order to write down the sentence "(*@\textcolor{red}{\{last sentence\}}@*)" I had to (*@\textcolor{blue}{research facts}@*)
\end{lstlisting}
\begin{lstlisting}[basicstyle=\footnotesize, escapeinside={(*@}{@*)},  caption={\textit{Guardrail hook.} Verbaliser tokens over which trigger is evaluated are shown in \textcolor{blue}{blue}. Placeholders for question-specific content are shown in \textcolor{red}{red}.}, label={lst:guardrail_trigger}]
Identify the sentences where information was failed to be found.
The sentence `So Legend Of The Sea Wolf came out first.' did not fail to find information
The sentence `There is no movie called Education Of A Prince.' failed to find information
The sentence `We need to find the birth years of the directors of both films.' did not fail to find information
The sentence `I couldn't find any information on Edward Bibring's wife or her occupation.' failed to find information
The sentence `Joe Jackson was nominated for a Grammy Award in 1979 for Best Male Pop Vocal Performance.' did not fail to find information
The sentence `Without any additional information, we cannot determine the birth order of Alexey Volkonsky and Wanderley Oliveira.' failed to find information
The sentence `Answer[Life Is Hot In Cracktown]' did not fail to find information
The sentence `(*@\textcolor{red}{\{last sentence\}}@*)' (*@\textcolor{blue}{failed to find}@*) information
\end{lstlisting}

\section{Prompts: programs} \label{sec:appendix-hook-prompts-program}

The below Listings show the prompt templates used by the programs in our experiments. Text in \textcolor{red}{red} denotes values which are placeholders for question-specific content either from the question itself or the rationale generated by the model. Text in black is part of the prompt template. 

\subsection{Calculator}

\begin{lstlisting}[basicstyle=\footnotesize, escapeinside={(*@}{@*)}, caption={Calculator program: \textit{Extract} any wrong calculations in the last sentence. Placeholders for question-specific content are shown in \textcolor{red}{red}.}, label={lst:program_calculator_extract}]
Question: Extract all of the calculations found in the sentence:
"That means he spent (13-4) * 9 = 9 * 9 = 81 dollars on ferris wheel tickets."
Write down each calculation on a seperate line. If there are no calculations respond with "no calculations". Write calculations in a python parsable format.
Answer:
(13-4) * 9 = 9 * 9
9 * 9 = 81

Question: Extract all of the calculations found in the sentence:
"Number of Years = $93 / $33852.825

Number of Years = 0.0027 years."
Write down each calculation on a seperate line. If there are no calculations respond with "no calculations". Write calculations in a python parsable format.
Answer:
93 / 33852.825 = 0.0027

Question: Extract all of the calculations found in the sentence:
"Therefore Bob has 3*(5+4)+9*(7+4)=126 toys."
Write down each calculation on a seperate line. If there are no calculations respond with "no calculations". Write calculations in a python parsable format.
Answer:
3*(5+4)+9*(7+4) = 126

Question: Extract all of the calculations found in the sentence:
"(*@\textcolor{red}{\{last sentence\}}@*)"
Write down each calculation on a seperate line. If there are no calculations respond with "no calculations". Write calculations in a python parsable format.
Answer:
\end{lstlisting}
\begin{lstlisting}[basicstyle=\footnotesize, escapeinside={(*@}{@*)}, caption={Calculator program: \textit{Format} extracted calculations to run in Python. Placeholders for question-specific content are shown in \textcolor{red}{red}.},  label={lst:program_calculator_format}]
Retranscribe the shown calculation as a single-line executable python script. Keep all numbers in the calculation unchanged and do not try and fix the mistakes. Remember to remove dollar signs and other units.

### Examples

Calculation : 13^9 = 117

Python script : ```13**9 == 117''' (but calculation is incorrect)

Calculation : 13,900dollarsx10books = 139,000

Python script : ```13900*10 == 139000'''

### Task

Calculation : (*@\textcolor{red}{\{calculation\}}@*)

Python script :
\end{lstlisting}
\begin{lstlisting}[basicstyle=\footnotesize, escapeinside={(*@}{@*)}, caption={Calculator program: \textit{Correct} any calculations in the last sentence. Placeholders for question-specific content are shown in \textcolor{red}{red}.},label={lst:program_calculator_correct}]
Q1: Rewrite "Text" so as to incorporate the information provided in "Facts". Make a minimal amount of changes and don't add any new information. If the facts are irrelevant you may ignore them and leave the text unchanged.
Text:
We know that each employee earns 7500 dollars which means that the 10 employee cummulatively earn 45000 dollars.

Facts:
7500*10 = 75000

Fixed text:
We know that each employee earns 7500 dollars which means that the 10 employee cummulatively earn 75000 dollars.

Q2: Rewrite "Text" so as to incorporate the information provided in "Facts". Make a minimal amount of changes and don't add any new information. If the facts are irrelevant you may ignore them and leave the text unchanged.
Text:
(*@\textcolor{red}{\{rationale\}}@*)

Facts:
(*@\textcolor{red}{\{corrected calculation\}}@*)

Fixed text:
\end{lstlisting}
\begin{lstlisting}[basicstyle=\footnotesize, escapeinside={(*@}{@*)}, caption={Calculator program: \textit{Trim} text after calculation. Placeholders for question-specific content are shown in \textcolor{red}{red}.},label={lst:program_trim}]
Question: Trim the following text (Original text) up until the end of the calculation (Calculation). If the calculation is irrelevant then leave the text as it is.
Original text: Adam bought 13 tickets and had 4 tickets left after going on the ferris wheel. That means he bought 13-4 = 9 tickets and spent 9 * 9 = 81 dollars on ferris wheel tickets.
Calculation:
13-4=9
Copied text:
 Adam bought 13 tickets and had 4 tickets left after going on the ferris wheel. That means he bought 13-4 = 9

Question: Trim the following text (Original text) up until the end of the calculation (Calculation). If the calculation is irrelevant then leave the text as it is.
Original text: In 4 weeks, she will eat <501894.75*28  = 14053053.0000000> dozen eggs. Therefore the answer is 14053053.0000000.
Calculation:
501894.75*28  = 14053053.0000000
Copied text:
 In 4 weeks, she will eat <501894.75*28  = 14053053.0000000>

Question: Trim the following text (Original text) up until the end of the calculation (Calculation). If the calculation is irrelevant then leave the text as it is.
Original text: (*@\textcolor{red}{\{rationale\}}@*)
Calculation:
(*@\textcolor{red}{\{corrected calculation\}}@*)
Copied text:
\end{lstlisting}

\subsection{Retriever}

\begin{lstlisting}[basicstyle=\footnotesize, escapeinside={(*@}{@*)}, caption={Retriever program: \textit{Generate} search queries. Placeholders for question-specific content are shown in \textcolor{red}{red}.}, label={lst:program_generate_search}]
Your task is to generate a set of 5 search queries for finding the wikipedia page containing the next relevant facts for progressing our answer. These keywords will be searched against to find the next wikipedia page and so should be:
1) comprehensive - they should contain all known knowledge. For example if we are searching for a movie director we should search for `name director' rather than `name'. This is helpful for filtering away similar pages.
2) targeted - they should not combine unrelated terms which would be unlikely to be found within a single wikipedia page. Instead they should be specific to one entity on wikipedia. For example they should not combine two unrelated movies.

### Examples

Question: `Are both The New Pornographers and Kings of Leon American rock bands?'
Rationale: Let's think step by step. We next need to search on Wikipedia for a page containing the next relevant facts for progressing our answer. Here are 5 variants on a set of keywords against which we could search to find the next wikipedia page:
`Kings of Leon'
`Kings of Leon band'
`New Pornographers'
`New Pornographers band'
`Kings of Leon country'

---

Question: `Who is the director of the 2003 film which has scenes in it filmed at the Quality Cafe in Los Angeles?'
Rationale: Let's think step by step. Films which have scenes filmed at the Quality Cafe in Los Angeles include Million Dollar Baby, Training Day, and Old School. We next need to search on Wikipedia for a page containing the next relevant facts for progressing our answer. Here are 5 variants on a set of keywords against which we could search to find the next wikipedia page:
`Million Dollar Baby'
`Million Dollar Baby movie year'
`Training day movie'
`Old School movie'
`Training day movie director'

---

Question: `In what year was the creator of the current arrangement of the "Simpson's Theme" born?'
Rationale: Let's think step by step. The creator of the Simpson's theme is Alf Clausen. We next need to search on Wikipedia for a page containing the next relevant facts for progressing our answer. Here are 5 variants on a set of keywords against which we could search to find the next wikipedia page:
`Alf Clausen age'
`Al Clausen composer birth year'
`Simpson's Alf Clausen born'
`Alf Clausen biography'
`Alf Clausen'

---

Question: `The Pineground Bridge formerly carried Depot Road over the Suncook River into a town with a population of what?'
Rationale: Let's think step by step. The Pineground Bridge formerly carried Depot Road over the Suncook River into Chichester, New Hampshire. We next need to search on Wikipedia for a page containing the next relevant facts for progressing our answer. Here are 5 variants on a set of keywords against which we could search to find the next wikipedia page:
`Chichester New Hampshire town population'
`Chichester New Hampshire'
`Chichester'
`Chichester New Hampshire census'
`Chicester census'

### Task

Question: `(*@\textcolor{red}{\{question\}}@*)'
Rationale: Let's think step by step. (*@\textcolor{red}{\{rationale\}}@*) We next need to search on Wikipedia for a page containing the next relevant facts for progressing our answer. Here are 5 variants on a set of keywords against which we could search to find the next wikipedia page:

\end{lstlisting}
\begin{lstlisting}[basicstyle=\footnotesize, escapeinside={(*@}{@*)}, caption={Retriever program: \textit{Re-generate} the last sentence. Placeholders for question-specific content are shown in \textcolor{red}{red}.},  label={lst:program_regenerate_last_sentence}]
Answer the question using the references made available to you. Not all references are relevant and you should use as few references as possible. Be careful to only use references:
1) Whose relevance is supported by the known information. Their content is already known to be relevant for the final answer.
2) Provide new information that will be key towards progressing our answer of the overarching goal.

### Examples

All references (randomly ordered non-comprehensive and sometimes off-topic):
[1] «The Theme | The Theme (Russian: Тема , "Tema " ) is a 1979 Soviet comedy film directed by Gleb Panfilov. It tells the story of an egotistical playwright who thinks of himself as an artist, but who allows the system to make him write conformist plays.»
[2] «The Simpsons Theme | "The Simpsons" Theme", also referred to as "The Simpsons" Main Title Theme" in album releases, is the theme music of the animated television series "The Simpsons". It plays during the opening sequence and was composed by Danny Elfman in 1989, after series creator Matt Groening approached him requesting a retro-style theme. The piece, which took 3 days, 2 hours, 48 minutes, and 19 seconds to create, has been noted by Elfman as the most popular of his career. The theme, as used for the opening sequence, was re-arranged during season 2, and the current arrangement by Alf Clausen was introduced at the beginning of the third season.»
[3] «Theming | Theming refers to "the use of an overarching theme...to create a holistic and integrated spatial organization of a consumer venue." It is also the subject of discourse, discussion, meditation and composition. In an overall sense, theming can be categorized under either "experience"; what an individual sees and feels in their current ‘themed environment’ or "décor"; which is utilized to make an individual remember something through the portrayal of the theme, whether it be generic or specific.»
[4] «The Simpsons | The Simpsons The Simpsons is an American animated sitcom created by Matt Groening for the Fox Broadcasting Company. The series is a satirical depiction of working-class life, epitomized by the Simpson family, which consists of Homer, Marge, Bart, Lisa, and Maggie. The show is set in the fictional town of Springfield and parodies American culture and society, television, and the human condition. The family was conceived by Groening shortly before a solicitation for a series of animated shorts with producer James L. Brooks.»
[5] «Alf Clausen | Alf Clausen Alf Heiberg Clausen (born March 28, 1941) is an American film and television composer. He is best known for his work scoring many episodes of "The Simpsons", of which he had been the sole composer between 1990 and 2017. Clausen has scored or orchestrated music for more than 30 films and television shows, including "Moonlighting", "The Naked Gun", "ALF" and "Ferris Bueller's Day Off". Clausen received an Honorary Doctorate of Music from the prestigious Berklee College of Music in 1996. Clausen was born in Minneapolis, Minnesota and raised in Jamestown, North Dakota. Clausen was interested in music from»

Question: """In what year was the creator of the current arrangement of the "Simpson's Theme" born?"""
Step-by-step answer: Let's think step by step. From reference [1] we learn that the creator of the current arrangement of the "Simpson's Theme" is Alf Clausen. From reference [5] we learn that Alf Clausen was born on March 28, 1941. So the final answer is March 28 1941.
Answer[March 28 1941]

---

All references (randomly ordered non-comprehensive and sometimes off-topic):
[1] «Training Day | Training Day is a 2001 American neo-noir crime thriller film directed by Antoine Fuqua, and written by David Ayer. Denzel Washington and Ethan Hawke star as two LAPD narcotics officers over a 24-hour period in the gang-ridden neighborhoods of the LAPD Rampart Division and South Central Los Angeles.»
[2] «Old School (film) | Old School is a 2003 American comedy film released by DreamWorks Pictures and The Montecito Picture Company and directed by Todd Phillips. The story was written by Court Crandall, and the film was written by Phillips and Scot Armstrong. The film stars Luke Wilson, Vince Vaughn, and Will Ferrell as three depressed thirty-somethings who seek to re-live their college days by starting a fraternity, and the tribulations they encounter in doing so. Since its release it has gained a massive cult following, since a lot of minor characters in the film went on to have huge careers such as Simon Helberg, Elisha Cuthbert, Rob Corddry and Artie Lange.»
[3] «Training Day (The Office) | "Training Day" is the twentieth episode of the seventh season of the American comedy television series "The Office" and the shows 146th episode overall. It originally aired on NBC on April 14, 2011. The episode was written by Daniel Chun and directed by Paul Lieberstein. This episode marks the first appearance of Deangelo Vickers (Will Ferrell) in the series.»
[4] «Quality Cafe (diner) | The Quality Cafe (also known as Quality Diner) is a now-defunct diner at 1236 West 7th Street in Los Angeles, California. The restaurant ceased to function as a diner in late 2006 but has appeared as a location featured in a number of Hollywood films, including "Training Day", "Old School", "Se7en", "Ghost World", "Gone in 60 Seconds", "The Stepfather", "What's Love Got to Do with It", "Sex and Death 101", and "Catch Me If You Can." It was also featured in Season 1 of the 2007 television series "Mad Men," in the episode "5G".»
[5] «Bad Day for Trains | Bad Day for Trains is the second studio album by Canadian country music singer/songwriter Patricia Conroy, and was released in 1992 by Warner Music Canada. The album was named Album of the Year by the Canadian Country Music Association in 1993.»
[6] «Million Dollar Baby (1941 film) | Million Dollar Baby is a 1941 romantic comedy film directed by Curtis Bernhardt. Released by Warner Bros., the film stars Priscilla Lane, Jeffrey Lynn, and Ronald Reagan, with May Robson and Lee Patrick.»

Question: """Who is the director of the 2003 film which has scenes in it filmed at the Quality Cafe in Los Angeles?"""
Step-by-step answer: Let's think step by step. From reference [4] we learn that films which have scenes filmed at the Quality Cafe in Los Angeles include "Training Day", "Old School", "Se7en", "Ghost World", "Gone in 60 Seconds", "The Stepfather", "What's Love Got to Do with It", "Sex and Death 101", and "Catch Me If You Can". From reference [2] we learn that Old School was released in 2003 and directed by Todd Phillips. So the final answer is Todd Phillips. 
Answer[Todd Phillips]

### Question

All references (randomly ordered non-comprehensive and sometimes off-topic):
(*@\textcolor{red}{\{references\}}@*)

Question: """(*@\textcolor{red}{\{question\}}@*)"""
Step-by-step answer: Let's think step by step.(*@\textcolor{red}{\{rationale\}}@*)
\end{lstlisting}

\section{Prompts: base model} \label{sec:appendix-hook-prompts-base}

The below Listings show the base prompt templates used by each method. Text in \textcolor{red}{red} denotes values which are placeholders for question-specific content either from the question itself or the rationale generated by the model. Text in black is part of the prompt template. Note that DSP constructs its own in-context examples.

\begin{lstlisting}[basicstyle=\footnotesize, escapeinside={(*@}{@*)}, caption={GSM8K base prompt for CoT and language hooks. Placeholders for question-specific content are shown in \textcolor{red}{red}.}, label={lst:gsm8k_prompt}]
Please answer each following question by firstly providing step by step rationale, and then providing the final answer as a single number at the end. Each step should make progress on the rationale little by little patiently, contain reasoning statement involving at most two entities or quantities, and contain at most one mathematical calculation. Each step should not contain calculations that are not explicitly stated, or complex reasoning involving multiple entities or quantities.

---

Question: Weng earns $12 an hour for babysitting. Yesterday, she just did 50 minutes of babysitting. How much did she earn?
Rationale: Let’s think step by step. Weng earns 12/60=0.2 dollar per minute. Weng worked for 50 minutes, so Weng earned 0.2*50=10 dollars.
Numerical final answer: 10

---

Question: James writes a 3-page letter to 2 different friends twice a week.  How many pages does he write a year?
Rationale: Let’s think step by step. James writes each friend 3*2=6 pages a week. So James writes 6*2=12 pages every week. So James writes 12*52=624 pages a year.
Numerical final answer: 624

---

Question: Mark has a garden with flowers. He planted plants of three different colors in it. Ten of them are yellow, and there are 80% more of those in purple. There are only 25% as many green flowers as there are yellow and purple flowers. How many flowers does Mark have in his garden?
Rationale: Let’s think step by step. There are 80/100*10=8 more purple flowers than yellow flowers. So there are 10+8=18 purple flowers. Purple and yellow flowers sum up to 10+18=28 flowers. So there are 25/100*28=7 green flowers. So in Mark's garden there are in total 28+7=35 flowers.
Numerical final answer: 35

---

Question: (*@\textcolor{red}{\{question\}}@*)
Rationale: Let’s think step by step.
\end{lstlisting}
\begin{lstlisting}[basicstyle=\footnotesize, escapeinside={(*@}{@*)}, caption={GSM8K base prompt for ReAct. Placeholders for question-specific content are shown in \textcolor{red}{red}.}, label={lst:gsm8k_react_prompt}]
Solve a question answering task with interleaving Thought, Action, Observation steps. Break down the problem into small substeps. Thought can reason about the current situation, and Action can be two types:
(1) Calculator[XopY], which at each call calculates a single operation as XopY and returns the answer. Calculator can also solve equations with a single free variable e.g. Calculator[3x=6].
(2) Finish[answer], which returns the answer as an integer and finishes the task.

Here are some examples.

---

Question: Weng earns $12 an hour for babysitting. Yesterday, she just did 50 minutes of babysitting. How much did she earn?
 Thought 1: I need to calculate how much money Weng earns per minute then multiply by the number of minutes Weng worked. Weng earns $12 an hour and an hour has 60 minutes.
Action 1: Calculator[12/60]
Observation 1: 12/60=0.2
Thought 2: Weng earns $0.2 per minute. I now need to calculate how much money Weng earned from working 50 minutes.
Action 2: Calculator[50*0.2]
Observation 2: 50*0.2=10
Thought 3: Weng earned $10 in total from working 50 minutes.
Action 3: Finish[10]

---

Question: James writes a 3-page letter to 2 different friends twice a week.  How many pages does he write a year?
 Thought 1: I need to calculate how many pages James writes a week then multiply by the number of weeks in a year.
Action 1: Calculator[3*2*2]
Observation 1: 3*2*2=12
Thought 2: James writes 12 pages a week. I now need to calculate how many pages we writes in a year. A year has 52 weeks.
Action 2: Calculator[12*52]
Observation 2: 12*52=624
Thought 3: James writes 624 pages per year.
Action 3: Finish[624]

---

Question: Mark has a garden with flowers. He planted plants of three different colors in it. Ten of them are yellow, and there are 80% more of those in purple. There are only 25% as many green flowers as there are yellow and purple flowers. How many flowers does Mark have in his garden?
 Thought 1: I need to calculate the number of purple flowers. Then I need to calculate the number of green flowers from the number of yellow and purple flowers. Finally I need to add up the number of yellow, purple and green flowers. Mark has 80% more purple flowers than yellow flowers, of which he has 10.
Action 1: Calculator[1.8*10]
Observation 1: 1.8*10=18
Thought 2: Mark has 18 purple flowers. I now need to calculate how many yellow and purple flowers Mark has.
Action 2: Calculator[10+18]
Observation 2: 10+18=28
Thought 3: Mark has 28 yellow and purple flowers. I now need to calculate how many green flowers Mark has.
Action 3: Calculator[0.25*28]
Observation 3: 0.25*28=7
Thought 4: Mark has 7 green flowers. I can now calculate how many flowers Mark has in total by adding up the number of yellow, purple and green flowers.
Action 4: Calculator[10+18+7]
Observation 4: 10+18+7=35
Thought 5: Mark has 35 flowers in total.
Action 5: Finish[35]

---

Question: (*@\textcolor{red}{\{question\}}@*)
\end{lstlisting}
\begin{lstlisting}[basicstyle=\footnotesize, escapeinside={(*@}{@*)},caption={GSM8K dataset base prompt for PAL. Placeholders for question-specific content are shown in \textcolor{red}{red}.}, label={lst:GSM8K_PAL_prompt}]
Question: Weng earns $12 an hour for babysitting. Yesterday, she just did 50 minutes of babysitting. How much did she earn?

# solution in Python:

 def solution():
    """Weng earns $12 an hour for babysitting. Yesterday, she just did 50 minutes of babysitting. How much did she earn?"""
    hourly_rate = 12
    minutes_worked = 50
    fraction_of_hour = minutes_worked / 60
    earnings = hourly_rate * fraction_of_hour
    result = earnings
    return result

---

Question: James writes a 3-page letter to 2 different friends twice a week.  How many pages does he write a year?

# solution in Python:

 def solution():
    """James writes a 3-page letter to 2 different friends twice a week. How many pages does he write a year?"""
    pages_per_letter = 3
    num_friends = 2
    letters_per_week = 2
    weeks_per_year = 52
    pages_per_year = pages_per_letter * num_friends * letters_per_week * weeks_per_year
    result = pages_per_year
    return result

---

Question: Mark has a garden with flowers. He planted plants of three different colors in it. Ten of them are yellow, and there are 80% more of those in purple. There are only 25% as many green flowers as there are yellow and purple flowers. How many flowers does Mark have in his garden?

# solution in Python:

 def solution():
    """Mark has a garden with flowers. He planted plants of three different colors in it. Ten of them are yellow, and there are 80% more of those in purple. There are only 25% as many green flowers as there are yellow and purple flowers. How many flowers does Mark have in his garden?"""
    yellow_flowers = 10
    purple_flowers = yellow_flowers * 1.8
    total_flowers = yellow_flowers + purple_flowers
    green_flowers = total_flowers * 0.25
    total_flowers += green_flowers
    result = total_flowers
    return result

---

Question: (*@\textcolor{red}{\{question\}}@*)

# solution in Python:
\end{lstlisting}
\begin{lstlisting}[basicstyle=\footnotesize, escapeinside={(*@}{@*)}, caption={HotpotQA base prompt for CoT and language hooks. Placeholders for question-specific content are shown in \textcolor{red}{red}.},  label={lst:hotpotqa_prompt}]
Answer the following two-hop trivia questions as best you. Always give an answer rather than saying that you don't know. None of the questions are trick questions.
Your response should comply to the following template:
Question: ${{the question needing answering}}
Rationale: ${{a step-by-step explanation, where the text is broken down into short sentences, which explicitly shows any facts, guesses and reasoning used to reach your final answer}}
Answer[${{your best guess as to the expected answer, obtained using your knowledge and any provided evidence}}]

---

Question: Who is the director of the 2003 film which has scenes in it filmed at the Quality Cafe in Los Angeles?
Rationale: Let’s think step by step. Films which have scenes filmed at the Quality Cafe in Los Angeles include Million Dollar Baby, Training Day, and Old School. Among them, Million Dollar Baby was released in 2005. Training Day was released in 2002. Old School was released in 2003. So the correct film is Old School. The director of Old School was Todd Phillips.
Answer[Todd Phillips]

---

Question: Are both The New Pornographers and Kings of Leon American rock bands?
Rationale: Let’s think step by step. The New Pornographers are a Canadian indie rock band. Kings of Leon is an American rock band. So one of them is not an American rock band.
Answer[no]

---

Question: In what year was the creator of the current arrangement of the "Simpson's Theme" born?'
Rationale: Let’s think step by step. The creator of the current arrangement of the "Simpson's Theme" is Alf Clausen. Alf Clausen was born on March 28, 1941.
Answer[1941]

---

Question: (*@\textcolor{red}{\{question\}}@*)
Rationale: Let’s think step by step.
\end{lstlisting}
\begin{lstlisting}[basicstyle=\footnotesize, escapeinside={(*@}{@*)}, caption={HotpotQA base prompt for ReAct. Placeholders for question-specific content are shown in \textcolor{red}{red}.}, label={lst:hotpotqa_react_prompt}]
Solve a question answering task with interleaving Thought, Action, Observation steps. Thought can reason about the current situation, and Action can be two types:
(1) Search[text], which searches and returns the first paragraph of Wikipedia articles based on the search term text. If no matches are found, it will return similar entities to search.
(2) Finish[answer], which returns the answer and finishes the task.

Here are some examples.

---

Question: Who is the director of the 2003 film which has scenes in it filmed at the Quality Cafe in Los Angeles?
 Thought 1: I need to search for Quality Cafe and find the director of the 2003 film which has scenes from there.
Action 1: Search[Quality Cafe]
Observation 1:
[1] «Quality Cafe | Quality Cafe is the name of two different former locations in Downtown Los Angeles, California.»
[2] «Quality Cafe (diner) | The Quality Cafe (also known as Quality Diner) is a now-defunct diner at 1236 West 7th Street in Los Angeles, California. The restaurant ceased to function as a diner in late 2006 but has appeared as a location featured in a number of Hollywood films, including "Training Day", "Old School", "Se7en", "Ghost World", "Gone in 60 Seconds", "The Stepfather", "What's Love Got to Do with It", "Sex and Death 101", and "Catch Me If You Can." It was also featured in Season 1 of the 2007 television series "Mad Men," in the episode "5G".»
[3] «Quality Cafe (jazz club) | Quality Cafe was a historical restaurant and jazz club located at 1143 East 12th Street near the corner of Central Avenue in Downtown Los Angeles. Quality Four, a jazz quartet founded by saxophonist Paul Howard and featuring young vibraphonist Lionel Hampton, was formed in 1924 to play at Quality Cafe. The band soon became Quality Quintet and then Quality Serenades, and was disbanded after a tour with Hazel Myers later in the same year. On June 7, 1924, the venue changed its name to Humming Bird Cafe and became "one of the hottest nightclubs in the area" under this name.»
Thought 2: There are several films filmed at the Quality Cafe (diner). I need to check which one is from 2003.
Action 2: Search[Training Day]
Observation 2:
[1] «Training Day | Training Day is a 2001 American neo-noir crime thriller film directed by Antoine Fuqua, and written by David Ayer. Denzel Washington and Ethan Hawke star as two LAPD narcotics officers over a 24-hour period in the gang-ridden neighborhoods of the LAPD Rampart Division and South Central Los Angeles.»
[2] «Training Day (The Office) | "Training Day" is the twentieth episode of the seventh season of the American comedy television series "The Office" and the shows 146th episode overall. It originally aired on NBC on April 14, 2011. The episode was written by Daniel Chun and directed by Paul Lieberstein. This episode marks the first appearance of Deangelo Vickers (Will Ferrell) in the series.»
[3] «Bad Day for Trains | Bad Day for Trains is the second studio album by Canadian country music singer/songwriter Patricia Conroy, and was released in 1992 by Warner Music Canada. The album was named Album of the Year by the Canadian Country Music Association in 1993.»
Thought 3: Training Day is from 2001. Let me search instead for Old School.
Action 3: Search[Old School]
Oberservation 3:
[1] «Old school | In slang, old school (or old skool) can refer to anything that is from an earlier era or anything that may be considered old-fashioned. The term is commonly used to suggest a high regard for something that has been shown to have lasting value or quality.»
[2] «Old Schools | The Old Schools are part of the University of Cambridge, in the centre of Cambridge, England. The Old Schools house the Cambridge University Offices, which form the main administration for the University.»
[3] «My Old School | "My Old School" is a single drawn from Steely Dan's 1973 album "Countdown to Ecstasy". It reached number 63 on the Billboard charts.»
Thought 4: That search didn't return a film. Let me refine my search.
Action 4: Search[Old School film]
Observation 4:
[1] «Old School (film) | Old School is a 2003 American comedy film released by DreamWorks Pictures and The Montecito Picture Company and directed by Todd Phillips. The story was written by Court Crandall, and the film was written by Phillips and Scot Armstrong. The film stars Luke Wilson, Vince Vaughn, and Will Ferrell as three depressed thirty-somethings who seek to re-live their college days by starting a fraternity, and the tribulations they encounter in doing so. Since its release it has gained a massive cult following, since a lot of minor characters in the film went on to have huge careers such as Simon Helberg, Elisha Cuthbert, Rob Corddry and Artie Lange.»
[2] «My Old School (film) | My Old School is a 2013 historical documentary created by Jason Allard and Scott Gabrielson.»
[3] «Old School New School (film) | Old School New School is an educational documentary film that examines the nature of creativity. It is produced by Steven Fischer and Diane Leigh Davison.»
Thought 5: Old School is a 2003 film. I need to find the director of it. The director of Old School is Todd Phillips.
Action 5: Finish[Todd Phillips]

---

Question: Are both The New Pornographers and Kings of Leon American rock bands?
 Thought 1: I need to search for The New Pornographers and Kings of Leon and check if they are both American rock bands.
Action 1: Search[The New Pornographers]
Observation 1:
[1] «The New Pornographers | The New Pornographers is a Canadian indie rock band formed in 1997 in Vancouver, British Columbia. Presented as a musical collective of singer-songwriters and musicians from multiple projects, the band has released seven studio albums to critical acclaim for their use of multiple vocalists and elements of power pop incorporated into their music.»
[2] «The Pornographers | The Pornographers is a 1966 Japanese film directed by Shohei Imamura and based on a novel ("Erogotoshitachi") by Akiyuki Nosaka. Its original Japanese title is ""Erogotoshitachi" yori Jinruigaku nyūmon" (「エロ事師たち」より 人類学入門 ), which means 'An introduction to anthropology through the pornographers'. It tells the story of porn film-maker Mr. Subuyan Ogata, whose business is under threat from thieves, the government, and his own family.»
[3] «The Pornographer | The Pornographer (French: Le Pornographe ) is a 2001 French-Canadian drama film written and directed by Bertrand Bonello who co-wrote the music score with Laurie Markovitch. The film features explicit sexual scenes by pornographic actress Ovidie. It won the FIPRESCI Prize (International Critics Week) at the 2001 Cannes Film Festival and was nominated for the Stockholm Film Festival for Bronze Horse.»
Thought 2: The New Pornographers are a Canadian indie rock band so I don't need to search for Kings of Leon.
Action 2: Finish[no]

---

Question: (*@\textcolor{red}{\{question\}}@*)
\end{lstlisting}
\begin{lstlisting}[basicstyle=\footnotesize, escapeinside={(*@}{@*)}, caption={Composite dataset base prompt for CoT and language hooks. Placeholders for question-specific content are shown in \textcolor{red}{red}.}, label={lst:composite_prompt}]
Answer the following questions as best you. Always give an answer rather than saying that you don't know. None of the questions are trick questions. Use short sentences.
Your response should comply to the following template:
Question: ${{the question needing answering}}
Rationale: ${{a step-by-step explanation, where the text is broken down into short sentences, which explicitly shows any facts, guesses and reasoning used to reach your final answer}}
So the answer to the first part multiplied by the answer to the second part is ${{a number representing your best guess as to the answer}}
Answer[${{the answer repeated}}]

---

Question: What is the answer to the first part multiplied by the answer to the second part?
First part: The Pineground Bridge formerly carried Depot Road over the Suncook River into a town with a population of what?
Second part: James creates a media empire.  He creates a movie for $2000.  Each DVD of the movie costs $6 to make.  He sells a DVD for 2.5 times that much.  He sells 500 DVDs a day for 5 days a week.  How much profit does he make in 20 weeks?
Rationale: Let’s think step by step. Firstly let's try to answer the first part step by step. The Pineground Bridge formerly carried Depot Road over the Suncook River into Chichester, New Hampshire. Chichester had a population of 2,523 people at the 2010 census. So the answer to the first part is 2523.
Then let's try to answer the second part step by step. James sales each DVD for 6*2.5=15 dollars. So James makes a profit of 15-6=9 dollars for each DVD. So each day James makes a profit of 9*500=4500 dollars. So James makes 4500*5=22500 dollars a week. So James makes 22500*20=450000 dollars in 20 weeks. Then after subtracting the cost of creating the movie, James has a profit of 450000-2000=448000 dollars. So the answer to the second part is 448000.
So the answer to the first part multiplied by the answer to the second part is 2523*448000=1130304000.
Answer[1130304000]

---

Question: What is the answer to the first part multiplied by the answer to the second part?
First part: James writes a 3-page letter to 2 different friends twice a week.  How many pages does he write a year?
Second part: Tura Beach, New South Wales is a suburb of a town that had what population at the 2016 census?
Rationale: Let’s think step by step. Firstly let's try to answer the first part step by step. James writes each friend 3*2=6 pages a week. So James writes 6*2=12 pages every week. So James writes 12*52=624 pages a year. So the answer to the first part is 624.
Then let's try to answer the second part step by step. Tura Beach is suburb of Merimbula, New South Wales, Australia. At the 2016 census, the population of Merimbula was 3,544. So the answer to the second part is 3544.
So the answer to the first part multiplied by the answer to the second part is 624*3544=2211456.
Answer[2211456]

---

Question: What is the answer to the first part multiplied by the answer to the second part?
First part: In what year did the submarine that sank the Tsushima Maru open to public tours?
Second part: Each bird eats 12 beetles per day, each snake eats 3 birds per day, and each jaguar eats 5 snakes per day. If there are 6 jaguars in a forest, how many beetles are eaten each day?
Rationale: Let’s think step by step. Firstly let's try to answer the first part step by step. Tsushima Maru was sunk by the submarine USS "Bowfin" during World War II. The submarine USS "Bowfin" has been open to public tours since 1981. So the answer to the first part is 1981.
Then let's try to answer the second part step by step. In the forest there are 6 jaguars eating in total 5*6=30 snakes per day. The 30 snakes eat in total 30*3=90 birds per day. The 90 birds eat in total 90*12=1080 beetles per day. So 1080 beetles are eaten in the forest each day. So the answer to the second part is 1080.
So the answer to the first part multiplied by the answer to the second part is 1981*1080=2139480.
Answer[2139480]

---

Question: (*@\textcolor{red}{\{question\}}@*)
Rationale: Let’s think step by step.
\end{lstlisting}
\begin{lstlisting}[basicstyle=\footnotesize, escapeinside={(*@}{@*)},caption={Composite dataset base prompt for ReAct. Placeholders for question-specific content are shown in \textcolor{red}{red}.}, label={lst:composite_react_prompt}]
Solve a question answering task with interleaving Thought, Action, Observation steps. Thought can reason about the current situation, and Action can be three types:
(1) Search[text], which searches and returns the first paragraph of Wikipedia articles based on the search term text. If no matches are found, it will return similar entities to search.
(2) Calculator[XopY], which calculates a single operation only as XopY and returns the answer. Calculator can also solve equations with a single free variable e.g. Calculator[3x=6].
(3) Finish[answer], which returns the answer and finishes the task.

Here are some examples.

---

Question: What is the answer to the first part multiplied by the answer to the second part?
First part: The Pineground Bridge formerly carried Depot Road over the Suncook River into a town with a population of what?
Second part: James creates a media empire.  He creates a movie for $2000.  Each DVD of the movie costs $6 to make.  He sells a DVD for 2.5 times that much.  He sells 500 DVDs a day for 5 days a week.  How much profit does he make in 20 weeks?
 Thought 1: I need to search for The Pineground Bridge.
Action 1: Search[The Pineground Bridge]
Observation 1:
[1] «Pineground Bridge | The Pineground Bridge, also known as the Depot Road Bridge or the Thunder Bridge, is a through-type lenticular truss bridge that formerly carried Depot Road over the Suncook River in Chichester, New Hampshire. The bridge was built in 1887 by the Berlin Iron Bridge Co., and is the only one of its type (and only one of for lenticular truss bridges of any type, as of 2004) in the state. It is 96 ft long and 16 ft wide, and rises 15 ft above the river. It is mounted on abutments made of unmortarted ashlar granite. The decking and rails are made of wood. The bridge has not undergone major alterations since its construction. Some of its stringers were replaced during the 1981-2 restoration, wooden parts were also replaced, and a number of decorative parts have been lost over the years.»
[2] «New Bridge (Dublin) | The New Bridge is a bridge over the River Dodder in Dublin. The bridge is also known as Herbert's Bridge or Lansdowne Bridge. The bridge is part of Lansdowne Road.»
[3] «Bridge therapy | Bridge therapy is therapy intended, in transportation metaphor, to serve as a figurative bridge to another stage of therapy or health, carrying a patient past a challenging period of some kind. There are various types of bridge therapy, such as bridge to transplant, bridge to candidacy, bridge to decision, bridge to recovery, and anticoagulation bridge (such as heparin bridge). Bridge therapy exists in contrast to destination therapy, which is the figurative destination rather than a bridge to something else.»
Thought 2: The Pineground Bridge formerly carried Depot Road over the Suncook River in Chichester, New Hampshire. I now need to search for Chichester, New Hampshire.
Action 2: Search[Chichester, New Hampshire]
Observation 2:
[1] «Chichester, New Hampshire | Chichester is a town in Merrimack County, New Hampshire, United States. The population was 2,523 at the 2010 census.»
[2] «Robert Trigges | Robert Trigges (by 1508-1550/1551), of St. Peter's, Chichester, Sussex and Chawton, Hampshire, was an English politician.»
[3] «Gabrielle Brune | Gabrielle Brune (12 February 1912 in Bournemouth, Hampshire – 18 January 2005 in Chichester, Sussex) was a British actress.»
Thought 3: Chichester has a population of 2,523 so the answer to the first part is 2523. I now need to solve the second part. I will work out the revenue per single DVD sale.
Action 3: Calculator[6*2.5]
Observtion 3: 6*2.5=15
Thought 4: I will now work out the profit per DVD.
Action 4: Calculator[15-6]
Observation 4: 15-6=9
Thought 5: So James makes $9 profit per DVD. I will now calculate his daily profit.
Action 5: Calculator[9*500]
Observation 5: 9*500=4500
Thought 6: James makes $4500 profit a day. I can now calculate his total profit from DVD sales.
Action 6: Calculator[4500*5*20]
Observation 6: 4500*5*20=450000
Thought 7: James makes $450000 in total. I now need to subtract the cost of movie creation to get his final profit.
Action 7: Calculator[450000-2000]
Observation 7: 450000-2000=448000
Thought 8: So James makes 448000 as his final profit in 20 weeks. So the answer to the second part is 448000. I can now calculate the answer to the first part multiplied by the answer to the second part.
Action 8: Calculator[2523*448000]
Observation 8: 2523*448000=1130304000
Thought 9: So the answer to the first part multiplied by the answer to the second part is 1130304000.
Action 9: Finish[1130304000]

---

Question: (*@\textcolor{red}{\{question\}}@*)
\end{lstlisting}

\twocolumn
\end{document}

%% file: schematic.tex
\begin{figure*}[t]
  \centering
\includegraphics[width=0.96\textwidth]{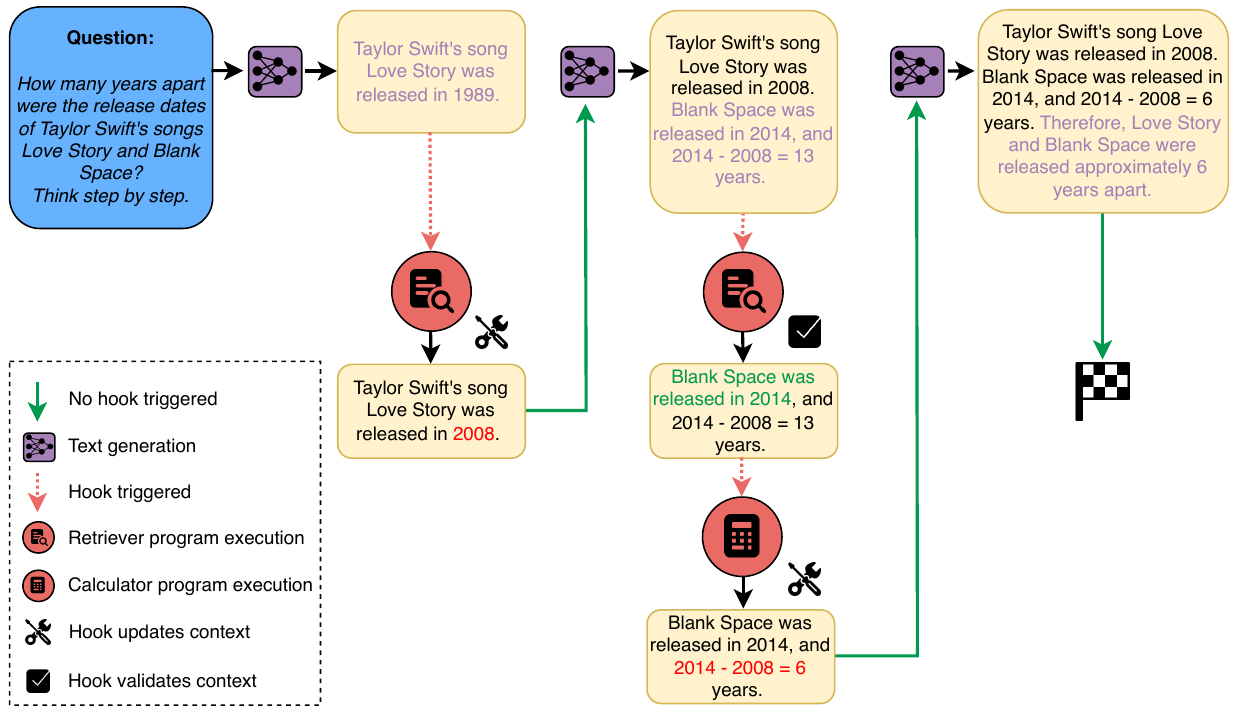}
\caption{Schematic of the language hook algorithm in a toy example. The base model generates text sentence by sentence. Each language hook has a trigger that monitors the advantage of running its program on the existing context. When a hook triggers, its associated program is executed (e.g.~knowledge retrieval or calculation check) and \textcolor[HTML]{FF0000}{modifies} (or \textcolor[HTML]{00994D}{not}) the existing context. When no hook triggers, the base model generates the \textcolor[HTML]{A680B8}{next sentence} of its response, conditioned on the existing context. This iterative process continues until a stopping condition is met.}
  \label{fig:schematic}
\end{figure*}


%% file: table_reasoning.tex
\begin{table*}[!th]
\centering
\small
\caption{Mathematical reasoning results quoted as exact-match (EM) scores.}
\vspace{2mm}
\begin{tabular}{ l @{\hspace{20pt}} c @{\hspace{20pt}}  c @{\hspace{20pt}} c @{\hspace{8pt}}}
\toprule
& \multicolumn{1}{c@{\hspace{20pt}}}{\textbf{GSM8K}}
& \multicolumn{1}{c@{\hspace{20pt}}}{\textbf{GSM-HARD}}
& \multicolumn{1}{c@{\hspace{20pt}}}{\textbf{GSM-HARD-filtered}}\\
& EM & EM & EM \\
\midrule
\textbf{CoT}            & 79.8 & 44.2 & 53.4 \\
\textbf{ReAct}          & 74.6 & 43.4 & 53.1 \\
\textbf{PAL}            & 78.8 & \textbf{55.8} & 57.7 \\
\midrule
\textbf{Language hooks} & \textbf{80.0} & 52.2 & \textbf{58.9} \\
\bottomrule
\end{tabular}%
\label{table:math-results}
\end{table*}


%% file: table_multihop.tex
\begin{table*}[tp]
\centering
\small
\caption{Multi-hop QA results in terms of EM and F1 scores.}
\vspace{2mm}
\begin{tabular}{ l @{\hspace{32pt}} cc @{\hspace{42pt}}  cc @{\hspace{42pt}} cc @{\hspace{0pt}}}
\toprule
& \multicolumn{2}{@{\hspace{11pt}}c@{\hspace{46pt}}}{\textbf{HotPotQA}}
& \multicolumn{2}{@{\hspace{6pt}}c@{\hspace{60pt}}}{\textbf{2WikiMultiHopQA}}\\
& \multicolumn{1}{@{\hspace{10pt}}c}{EM} & \multicolumn{1}{@{\hspace{12pt}}l}{F1} & \multicolumn{1}{@{\hspace{22pt}}c}{EM} & \multicolumn{1}{@{\hspace{12pt}}l}{F1} \\
\midrule

\textbf{CoT} & \multicolumn{1}{@{\hspace{10pt}}c}{28.0} & \multicolumn{1}{@{\hspace{8pt}}l}{40.9} & \multicolumn{1}{@{\hspace{22pt}}c}{28.8} & \multicolumn{1}{@{\hspace{8pt}}l}{36.4}\\

\textbf{ReAct} & \multicolumn{1}{@{\hspace{10pt}}c}{28.8} & \multicolumn{1}{@{\hspace{8pt}}l}{41.2} & \multicolumn{1}{@{\hspace{22pt}}c}{26.2} & \multicolumn{1}{@{\hspace{8pt}}l}{35.0} \\

\textbf{ReAct  $\rightarrow$ CoT} & \multicolumn{1}{@{\hspace{10pt}}c}{35.4} & \multicolumn{1}{@{\hspace{8pt}}l}{50.2} & \multicolumn{1}{@{\hspace{22pt}}c}{32.6} & \multicolumn{1}{@{\hspace{8pt}}l}{44.4} \\

\textbf{DSP with SC} & \multicolumn{1}{@{\hspace{10pt}}c}{38.6} & \multicolumn{1}{@{\hspace{8pt}}l}{\textbf{53.9}} & \multicolumn{1}{@{\hspace{22pt}}c}{36.4} & \multicolumn{1}{@{\hspace{8pt}}l}{48.3} \\

\midrule
\textbf{Language hooks} & \multicolumn{1}{@{\hspace{10pt}}c}{\textbf{39.0}} & \multicolumn{1}{@{\hspace{8pt}}l}{53.7} & \multicolumn{1}{@{\hspace{22pt}}c}{\textbf{47.0}} & \multicolumn{1}{@{\hspace{8pt}}l}{\textbf{60.0}} \\
\bottomrule
\end{tabular}%
\label{table:multihop-results}
\end{table*}

%% file: table_guardrail.tex

\begin{table}[t]
\small
\centering
\captionof{table}{Language hooks provide effective external validation on HotpotQA (HPQA) and 2WikiMultihopQA (2Wiki). See Figure \ref{fig:guardrail-ablation} for setup.}
\vspace{2mm}
\begin{tabular}{ l @{\hspace{12pt}} cc}
\toprule
& \textbf{HPQA} & \textbf{2Wiki} \\
& F1 & F1 \\
\midrule
$\mathbf{S_1}$ [base model] & 46.6 & 43.1 \\
$\mathbf{S_3}$ [guardrail hook only]  & 29.8 & 26.0 \\
$\mathbf{S_3}$ [guardrail \& retriever] & 45.0 & 52.2 \\
\bottomrule
\end{tabular}
\label{table:ablation-guardrail}
\end{table}

%% file: table_multihook.tex
\begin{table}[t]
\centering
\small
\captionof{table}{Composite QA results.}
\vspace{2mm}
\begin{tabular}{ l @{\hspace{12pt}} c}
\toprule
& \multicolumn{1}{c@{\hspace{12pt}}}{\textbf{HotpotQA$\times$GSM8K}} \\
& EM \\
\midrule
\textbf{CoT}            & 19.0 \\
\textbf{ReAct}          & 30.3 \\
\textbf{ReAct  $\rightarrow$ CoT} & 31.0 \\
\midrule
\textbf{Language hooks} & \textbf{46.3} \\
\bottomrule
\end{tabular}
\label{table:multihook-results}
\end{table}

%% file: table_ablation_composite_simple.tex
\begin{table}[t]
\centering
\small
\captionof{table}{Measuring the ability of ReAct and language hooks (LH) to select tools and compose responses on the composite QA task. The expected composite performance is calculated from \textbf{(A1)} $\times$ \textbf{(A2)} question pairs.}
\vspace{2mm}
\begin{tabular}{ l @{\hspace{12pt}} cc}
\toprule 
& \multicolumn{1}{@{\hspace{0pt}}l}{\textbf{ReAct}}
& \multicolumn{1}{@{\hspace{11pt}}l}{\textbf{LH}} \\
& EM & EM \\
\midrule
\textbf{(A1)} GSM8K sub-Q [calculator]   & 77.3 & 81.3 \\
\textbf{(A2)} HotpotQA sub-Q [retriever] & 53.7 & 69.3 \\
Expected composite performance           & 44.0 & 57.0 \\
\midrule
Composite Q (Table \ref{table:multihook-results})                 & 30.3 & 46.3 \\
Relative performance drop            & 31.1\% & 18.8\% \\
\bottomrule
\end{tabular}
\label{table:ablation-composite-simple}
\end{table}

%% file: table_ablation_composite.tex
\begin{table*}[t]
\centering
\small
\caption{\textbf{Ablation: tool selection.} Measuring the ability of methods to select tools on the individual sub-questions for HotpotQA$\times$GSM8K. We also show CoT performance when answering the sub-questions individually though this doesn't use any tools. Expected composite performance is calculated from question pairs in the composite QA dataset.}
\vspace{2mm}
\begin{tabular}{ l @{\hspace{20pt}} c @{\hspace{20pt}}  c @{\hspace{20pt}} c @{\hspace{8pt}}}
\toprule
& \multicolumn{1}{@{\hspace{4pt}}l}{\textbf{CoT}}
& \multicolumn{1}{@{\hspace{0pt}}l}{\textbf{ReAct}}
& \multicolumn{1}{@{\hspace{0pt}}l}{\textbf{Language hooks}}\\
& EM & EM & EM \\
\midrule
\textbf{(A1)} GSM8K sub-Q - calculator only (no tool for CoT) & 79.3 & 77.3 & 81.3 \\
\textbf{(A2)} HotpotQA sub-Q - retriever only (no tool for CoT) & 46.7 & 53.7 & 69.3 \\
(\textbf{A1} $\times$ \textbf{A2}) Expected composite performance & 34.3 & 44.0 & 57.0 \\
\midrule
\textbf{(B1)} GSM8K sub-Q - multi-tool                          & -    & 74.3 & 78.3 \\
\textbf{(B2)} HotpotQA sub-Q  - multi-tool                      & -    & 58.7 & 66.7 \\
(\textbf{B1} $\times$ \textbf{B2}) Expected composite performance  & -    & 44.3 & 52.7 \\
\midrule
\textbf{(C)} Composite Q - multi-tool (per Table \ref{table:multihook-results}) & 19.0 &  30.3 & 46.3 \\
Performance drop (\textbf{(C)} relative to \textbf{A1} $\times$  \textbf{A2}) & 44.6\% & 31.1\% & 18.8\% \\
Performance drop (\textbf{(C)}  relative to \textbf{B1} $\times$  \textbf{B2}) & -      & 31.6\% & 12.1\% \\
\bottomrule
\end{tabular}%
\label{table:ablation-composite-full}
\end{table*}

%% file: table_ablation_composite_deconstructed.tex
\begin{table*}[t]
\centering
\small
\caption{\textbf{Decomposing composite QA performance.} Here we breakdown the composite QA performance from Table \ref{table:multihook-results}. $P(X)$ represents the probability of correctly answering a question of type $X$ whilst reasoning where $X \in \{\text{GSM8K}, \text{HotpotQA}, \text{composite}\}$. $P((X, Y) \text{ in final})$ represents both sub-question answers being correctly inserted into the model's final calculation. Note that $P(\text{composite})$ is the same as shown in Table \ref{table:multihook-results}.}
\vspace{2mm}
\begin{tabular}{ l @{\hspace{20pt}} c @{\hspace{20pt}}  c @{\hspace{20pt}} c @{\hspace{8pt}}}
\toprule
& \multicolumn{1}{@{\hspace{2pt}}l}{\textbf{CoT}}
& \multicolumn{1}{@{\hspace{-1pt}}l}{\textbf{ReAct}}
& \multicolumn{1}{@{\hspace{0pt}}l}{\textbf{Language hooks}}\\
\midrule
$P$(GSM8K)                       & 0.787 & 0.643 & 0.803 \\
$P$(HotpotQA)                    & 0.477 & 0.573 & 0.623 \\
$P$(GSM8K, HotpotQA)             & 0.367 & 0.390 & 0.500 \\
\midrule
\textbf{(E)} $P$((GSM8K, HotpotQA) in final)    & 0.353 & 0.310 & 0.473 \\
\textbf{(F)} $P$(composite | (GSM8k, HotpotQA) in final) & 0.538 & 0.978 & 0.979 \\
\midrule
\textbf{(E $\times$ F)} $P$(composite)                   & 0.190 & 0.303 & 0.463 \\
\bottomrule
\end{tabular}%
\label{table:ablation-composite-deconstructed}
\end{table*}
